%% file: 0_Main.tex
\documentclass[letterpaper]{article} % DO NOT CHANGE THIS
\usepackage[preprint]{aaai2027}

\usepackage[hyphens]{url}  % DO NOT CHANGE THIS
\usepackage{graphicx} % DO NOT CHANGE THIS
\usepackage{natbib}  % DO NOT CHANGE THIS AND DO NOT ADD ANY OPTIONS TO IT
\usepackage{caption} % DO NOT CHANGE THIS AND DO NOT ADD ANY OPTIONS TO IT
\usepackage{algorithm}
\usepackage{algorithmic}

\usepackage{booktabs}
\usepackage{multirow}
\usepackage{amsmath}
\usepackage{amsthm}
\usepackage{amssymb}

\usepackage{color,xcolor}
\usepackage{makecell}

\usepackage{newfloat}
\usepackage{listings}
\DeclareCaptionStyle{ruled}{labelfont=normalfont,labelsep=colon,strut=off} % DO NOT CHANGE THIS
\floatstyle{ruled}
\newfloat{listing}{tb}{lst}{}
\floatname{listing}{Listing}

\usepackage{booktabs}
\usepackage{pifont}    % \ding

\newcommand{\cmark}{\ding{51}}%
\newcommand{\xmark}{\ding{55}}%
\newcommand{\valstd}[2]{$#1${\tiny $\pm #2$}}

\newtheorem{theorem}{Theorem}
\newtheorem{corollary}{Corollary}

\definecolor{mygreen}{HTML}{2CA25F}
\definecolor{myred}{HTML}{d62728}

\title{Marginal-Preserving Coupled Diffusion Composition for Long-Horizon Planning}

\author{
    Hengrui Zhang\textsuperscript{\rm 1},
    Qiyu Wang\textsuperscript{\rm 2},
    Yuhu Cheng\textsuperscript{\rm 1},
    Brooks Paige\textsuperscript{\rm 2},
    Xuesong Wang\textsuperscript{\rm 1}\thanks{Corresponding author.}
}

\affiliations{
    \textsuperscript{\rm 1}School of Information and Control Engineering, China University of Mining and Technology\\
    \textsuperscript{\rm 2}Center for Artificial Intelligence, University College London, London \\
    hengruizhang@cumt.edu.cn, chengyuhu@cumt.edu.cn, wangxuesong@cumt.edu.cn\\
    qiyu.wang.23@ucl.ac.uk, b.paige@ucl.ac.uk\\
}

\begin{document}

\maketitle

\begin{abstract}
Compositional diffusion planners extend short-horizon trajectory models to long-horizon goal-conditioned planning by generating and stitching multiple local trajectory chunks. However, in multimodal environments, local plausibility is not sufficient for global consistency: independently initialized chunks may denoise toward incompatible overlap modes, producing plans that are individually feasible but incoherent after stitching. Existing remedies typically add search, guidance, pruning, or graph-based structure at inference time, increasing computational cost or relying on task-specific machinery. We instead investigate graph-structured dependence among chunk-wise initial diffusion noises as a lightweight coordination mechanism for compositional trajectory planning.
We propose \textbf{M}arginal-Preserving \textbf{C}oupled \textbf{D}iffusion \textbf{C}omposition (MCDC), a training-free inference method for pretrained compositional diffusion planners. Instead of sampling chunk noises independently, MCDC constructs coupled Gaussian initializations whose per-chunk marginals remain exactly standard Gaussian, as required by the pretrained denoiser, while neighboring chunks receive positively correlated noise components. This preserves the standard-Gaussian initialization marginal of each chunk while biasing adjacent chunks toward compatible overlap-mode choices. We prove exact marginal preservation and show, under a simple multimodal overlap model, that positive coupling increases the probability of compatible mode selection. On OGBench long-horizon stitch tasks, MCDC consistently improves over independent compositional diffusion sampling while retaining nearly the same inference time, and achieves a favorable success--cost trade-off compared with search-based composition methods.
\end{abstract}


\input{1_Introduction}

\input{2_Related_Works}

\input{3_Preliminaries}

\input{4_Motivation}

\input{5_Method}

\input{6_Experiments}

\input{7_Conclusion}

\clearpage
\bibliography{reference}

\clearpage

\clearpage
\onecolumn

\setcounter{secnumdepth}{2}
\appendix

\renewcommand{\thesubsection}{\thesection.\arabic{subsection}}
\renewcommand{\thesubsubsection}{\thesubsection.\arabic{subsubsection}}

\begin{center}
    {\LARGE\bfseries Technical Supplement}
\end{center}

\input{8_Appendix}

\end{document}

%% file: 1_Introduction.tex
\section{Introduction}
Long-horizon goal-conditioned reinforcement learning requires an agent to reach distant goals by composing multiple locally feasible behaviors \cite{GSC, CompDiffuser, RCD}. In the offline setting, this problem is especially challenging because the dataset is fixed, successful long-horizon trajectories are scarce, and test-time start--goal pairs often do not appear as complete trajectories in the data \cite{OGBench, HIQL, DSP}. A practical planner must therefore reuse short-horizon experience while remaining coherent over much longer horizons.

 Diffusion models offer a natural planning-as-inference framework for this setting \cite{Diffuser, AdaptDiffuser, SkillDiffuser}. Trained on offline trajectories, they can generate dynamically plausible goal-conditioned behaviors and have shown strong performance in trajectory planning \cite{Diffuser, DD}. However, monolithic diffusion planners remain constrained by the temporal horizon and coverage of the training data; directly sampling long trajectories from short-horizon data can move the sampler outside the learned distribution \cite{CompositionalDiffusion, CinDM}. Compositional diffusion planners mitigate this limitation by decomposing a long plan into short-horizon trajectory segments and stitching them through overlap regions or neighboring conditions at inference time. Prior work \cite{GSC, CompDiffuser, CDGS, RCD} has shown that such generative stitching enables short-horizon diffusion models to generalize to longer goal-reaching problems.

The key difficulty is that compositional consistency is not only a geometric constraint. In multimodal environments, the same local region may support several valid passages, detours, or intermediate subgoals. Two neighboring chunks can therefore be individually plausible while committing to incompatible modes in their shared overlap. For example, one chunk may approach an overlap through the upper corridor while the next leaves through the lower corridor. Each segment can look valid in isolation, yet their stitched trajectory becomes globally incoherent, producing discontinuities, wall-crossing connections, or low-density transitions. This failure is not necessarily caused by a weak local denoiser; it can arise from insufficient coordination among otherwise valid local samples \cite{CDGS, XDiffuser, RCD}.

Existing approaches typically address this issue by introducing additional coordination machinery at inference time. Search-based methods sample or prune multiple candidate compositions \cite{CDGS}. Guidance-based methods optimize auxiliary consistency or value objectives during denoising \cite{RCD, CVP}. Graph- or waypoint-based methods impose a global structure before local trajectory generation \cite{XDiffuser}. While effective, these strategies introduce additional inference-time machinery beyond the base compositional sampler, including search procedures, guidance objectives, or task-specific structural priors.

We instead revisit a basic sampling assumption in compositional diffusion: different trajectory segments are typically initialized with \textit{independent} diffusion noise. This independence is not required by the pretrained denoiser; it is a choice made by the inference sampler. In multimodal overlap regions, independent random sources can drive neighboring chunks toward independently selected modes, increasing the chance of incompatible stitching. Coupling the random sources of neighboring chunks, while preserving the standard Gaussian marginal expected by each local denoiser, offers a way to bias compositional sampling toward compatible overlap modes without changing the pretrained model.

Motivated by this observation, we propose \textbf{M}arginal-Preserving \textbf{C}oupled \textbf{D}iffusion \textbf{C}omposition (MCDC), a training-free inference method for compositional diffusion planning. MCDC does not retrain the diffusion model, add a value function, or introduce a learned verifier. It only reconstructs the Gaussian noise used by multiple chunks at inference time, preserving the local standard Gaussian marginal of each chunk while inducing controllable local or hybrid global-local correlations across neighboring chunks. Thus, MCDC preserves each chunk’s initial Gaussian marginal, while altering cross-chunk dependence.
The contributions are summarized as follows:
\begin{itemize}
    \item We identify independent chunk-noise initialization as a sampler-level source of mode conflict in compositional diffusion planning.
    \item We propose MCDC, a drop-in coupled-noise initialization method that preserves each chunk's Gaussian marginal while coordinating neighboring chunks through controlled cross-chunk dependence.
    \item We analyze why positive coupling improves overlap-mode compatibility under a multimodal basin model.
    \item We show on OGBench long-horizon goal-reaching tasks that MCDC improves compositional diffusion planning with negligible inference overhead and without retraining, guidance optimization, or candidate search.
    % \item We characterize a sampler-level failure mode in compositional diffusion planning: independent chunk-noise initialization can induce independent mode choices in multimodal overlaps, leading to locally feasible but globally incompatible stitching.
    % \item We propose MCDC, a training-free marginal-preserving coupled-noise initialization method that structures Gaussian dependence over the chunk adjacency graph of a compositional planner, while leaving the pretrained denoiser, reverse process, and stitching rule unchanged.
    % \item We prove that MCDC preserves local Gaussian diffusion marginals and, under a multimodal overlap-basin model, increases the probability of compatible mode selection; experiments on OGBench long-horizon offline goal-reaching tasks validate its effectiveness.
\end{itemize}

%% file: 2_Related_Works.tex
\section{Related Work}

\paragraph{Offline goal-conditioned reinforcement learning.}
Offline goal-conditioned reinforcement learning (GCRL) aims to learn goal-reaching behavior from fixed datasets without additional environment interaction \cite{OGBench, HIQL, PCDT}. Representative methods relabel future states as goals \cite{HER, CHER}, train goal-conditioned policies through supervised learning \cite{GCSL, WGCSL, Goplan}, estimate value or advantage functions for offline goal reaching \cite{Pi-HIQL, OTA, SAW}, or learn state--goal representations that encode reachability \cite{CRL, QRL, GoFAR, SVL, LAVL}. Generative planners form another family: they model trajectories, subgoals, skills, or policies as conditional generative distributions and use the generated outputs for offline decision making \cite{Diffuser, DD, AdaptDiffuser, SkillDiffuser, DiffusionQL, FQL, CFGRL}. These works show that conditional generative modeling can capture multimodal behaviors from offline data and support flexible planning. MCDC follows this generative-planning perspective, but shifts the focus from learning a new decision model to coordinating pretrained local diffusion planners at inference time.

\paragraph{Compositional diffusion models for trajectory planning.}
Diffusion and related continuous-time generative models \cite{DDPM, DDIM, Score_SDE, FM, RF} provide flexible tools for modeling multimodal trajectory distributions. In offline decision making and trajectory planning, such models have been used to generate trajectories, policies, or value-improved behaviors from fixed datasets \cite{Diffuser, HD, HDMI, DTAMP}. For long-horizon goal reaching, compositional diffusion planners further decompose a global plan into short-horizon chunks and compose them through overlap regions or neighboring conditions \cite{CompositionalDiffusion, GSC, CompDiffuser}. This improves horizon generalization, but introduces a compositional consistency problem: locally plausible chunks may still select incompatible modes in their shared overlaps. MCDC targets this consistency failure at the sampler-initialization level, coupling neighboring chunk noises while preserving each chunk's standard Gaussian marginal.

\paragraph{Inference-time refinement.}
Inference-time refinement improves pretrained generative planners by changing how samples are generated or selected at test time \cite{TDP}. Existing methods search or prune candidate plans and denoising branches \cite{CDGS, MCTD, FastMCTD, GTS}, guide sampling with auxiliary consistency or prior objectives \cite{RCD, PG}, or impose global scaffolds such as waypoint and graph structures \cite{XDiffuser, CVP}. MCDC is orthogonal to these refinements: it changes only the joint initialization of chunk noises, preserving each local Gaussian marginal while introducing cross-chunk correlations that encourage overlap compatibility.

%% file: 3_Preliminaries.tex
\section{Preliminaries}
\begin{figure*}[t]
    \centering
    \includegraphics[width=\linewidth]{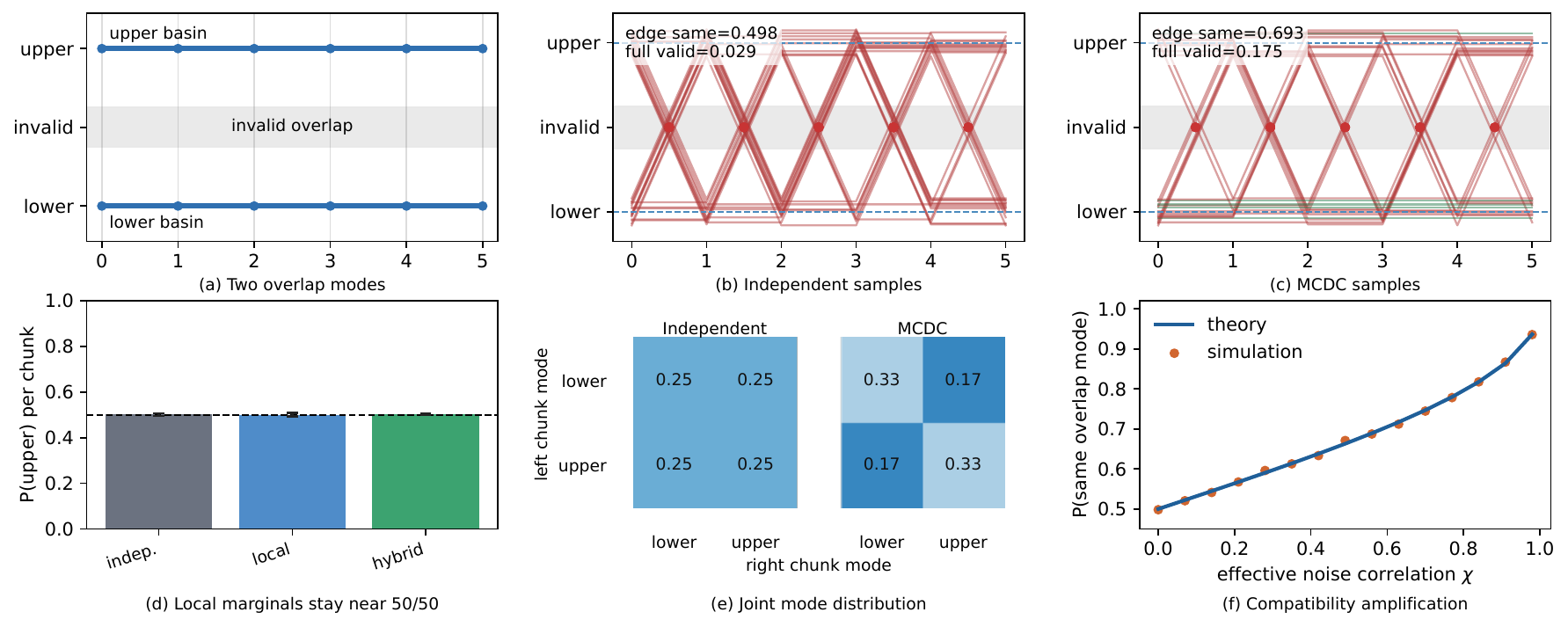}
    \vspace{-0.3cm}
    \caption{
    \textbf{Motivating corridor-basin toy example.}
    \textbf{(a)} Two valid overlap modes are separated by an invalid region.
    \textbf{(b)} Independent chunk sampling often switches modes across overlaps, producing invalid (\textbf{red}) composed trajectories.
    \textbf{(c)} MCDC promotes same-mode stitching and produces valid (\textbf{green}) composed trajectories.
    \textbf{(d)} Local mode marginals remain close to 50/50 under independent, local, and hybrid sampling.
    \textbf{(e)} Coupled noise shifts joint mode mass from incompatible off-diagonal pairs to compatible diagonal pairs.
    \textbf{(f)} The Gaussian sign model and simulation show that positive noise correlation amplifies overlap compatibility.
    Toy-setting details are provided in Appendix~\ref{app:toy_setting}.
    }
    \label{fig:toy}
\end{figure*}

\paragraph{Problem Setting.}
We consider offline goal-conditioned reinforcement learning with state space $\mathcal{S}$, action space $\mathcal{A}$, transition dynamics $p(s' \mid s,a)$, and goal space $\mathcal{G}$. The agent is given a fixed dataset $\mathcal{D}$ of trajectories collected without access to the test-time start--goal pairs. Following generative planning methods for offline GCRL, we plan state trajectories $\tau = (s_0, s_1, \ldots, s_H)$ conditioned on an initial state $s_0$ and a goal $g \in \mathcal{G}$. The generated state trajectory is then executed through an inverse dynamics model or a low-level controller, separating long-horizon planning in state space from short-horizon action execution.

\paragraph{Diffusion models for trajectory generation.}
Diffusion planners model trajectories with a conditional generative model. Let $\tau^0$ denote a clean trajectory segment from the offline dataset and $\tau^t$ its noisy version at diffusion step $t$. The forward process adds Gaussian noise:
\begin{align}
    q(\tau^t \mid \tau^0)
    =
    \mathcal{N}\!\left(
    \sqrt{\bar{\alpha}_t}\tau^0,
    (1-\bar{\alpha}_t)I
    \right),
\end{align}
or equivalently,
\begin{align}
    \tau^t
    =
    \sqrt{\bar{\alpha}_t}\tau^0
    +
    \sqrt{1-\bar{\alpha}_t}\epsilon,
    \qquad
    \epsilon\sim\mathcal{N}(0,I).
\end{align}
A denoising network $\epsilon_\theta(\tau^t,t,c)$ predicts the injected noise under condition $c$, such as start and goal constraints. At inference time, sampling starts from $\tau^T\sim\mathcal{N}(0,I)$ and iteratively applies
\begin{align}
    p_\theta(\tau^{t-1}\mid \tau^t,c)
    =
    \mathcal{N}\!\left(
    \mu_\theta(\tau^t,t,c),
    \sigma_t^2 I
    \right),
\end{align}
where $\mu_\theta$ is computed from $\epsilon_\theta(\tau^t,t,c)$. The final denoised sample $\tau^0$ is used as the planned trajectory.

\paragraph{Compositional diffusion planning.}
For long-horizon tasks, directly generating a single trajectory may exceed the horizon and coverage of the offline data. Compositional diffusion planners address this by decomposing a global plan into $K$ overlapping short-horizon chunks,
\begin{align}
    \boldsymbol{\tau}=(\tau_1,\tau_2,\ldots,\tau_K).
\end{align}
A common chunk-based formulation represents the long-horizon trajectory distribution through local conditional chunk distributions:
\begin{align}
    p_\theta(\boldsymbol{\tau}\mid s_0,g)
    \propto
    &p_\theta(\tau_1\mid s_0,\tau_2)
    p_\theta(\tau_K\mid \tau_{K-1},g) \notag \\
    & \cdot \prod_{k=2}^{K-1}
    p_\theta(\tau_k\mid \tau_{k-1},\tau_{k+1}),
\end{align}
where $s_0$ and $g$ denote the start and goal conditions. In practice, each chunk is initialized with Gaussian noise,
\begin{align}
    \tau_k^T=\epsilon_k^{\mathrm{ind}},
    \qquad
    \epsilon_k^{\mathrm{ind}}\sim\mathcal{N}(0,I),
\end{align}
and denoised with local compositional conditions:
\begin{align}
    p_\theta(\tau_k^{t-1}\mid \tau_k^t,c_k^t)
    =
    \mathcal{N}\!\left(
    \mu_\theta(\tau_k^t,t,c_k^t),
    \sigma_t^2 I
    \right).
\end{align}
Here $c_k^t$ denotes planner-specific conditioning, such as start--goal constraints, neighboring noisy chunks, or overlap-based score composition. The denoised chunks are then merged through their overlap regions to form the final trajectory.

%% file: 4_Motivation.tex
\section{Motivation: Mode Conflict from Independent Chunk Sampling}

\begin{figure*}[t]
    \centering
    \includegraphics[width=0.91\linewidth]{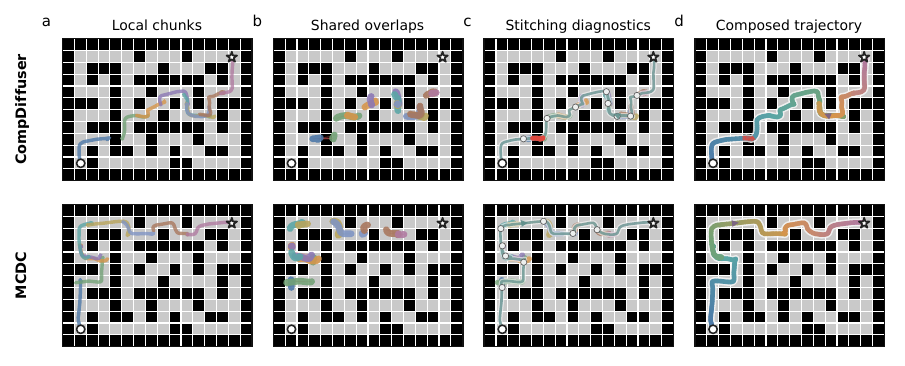}
    \vspace{-0.1cm}
    \caption{
    \textbf{Mode conflict in \texttt{AntMaze-Giant-Stitch} task 1.}
    \textbf{(a)} Local chunks generated by CompDiffuser and MCDC.
    \textbf{(b)} Shared overlap regions between neighboring chunks.
    \textbf{(c)} Stitching diagnostics, where red marks incompatible wall-crossing segments.
    \textbf{(d)} Final composed trajectories; MCDC yields a coherent composition in this example.
    }
    \label{fig:comparison_visualization_task1}
\end{figure*}

Compositional diffusion planning requires both local feasibility and cross-chunk compatibility. Each generated chunk should be plausible under the pretrained local denoiser, but neighboring chunks must also agree on how their shared overlap is used. In multimodal environments, this second requirement is nontrivial: the same overlap region may admit multiple feasible passages, detours, or intermediate states. Therefore, locally plausible chunks can still become globally inconsistent after stitching.

Figure~\ref{fig:toy} illustrates this failure mode with a two-basin overlap example. Each chunk can choose either the upper or lower basin, and both choices are locally feasible. However, a valid composition requires adjacent chunks to select compatible basins in the shared overlap. If neighboring chunks select different basins, the stitched trajectory may contain wall-crossing connections, discontinuities, or low-density transitions. The issue is therefore not necessarily that the local denoiser generates invalid chunks; rather, independent random initializations can allow neighboring denoising chains to commit to different valid overlap modes.

To make this intuition explicit, consider a simplified two-basin overlap abstraction. Let $b_k \in \{-1,+1\}$ denote the basin selected by chunk $k$ in the overlap region. If two neighboring chunks choose their basins independently with balanced marginals, then
\begin{align}
    \Pr(b_k=b_{k+1})=\frac{1}{2}.
\end{align}

Now suppose the basin choice is determined by an early denoising coordinate, $b_k=\operatorname{sign}(\xi_k)$, where $(\xi_k,\xi_{k+1})$ are jointly Gaussian with standard Gaussian marginals and effective correlation $\chi$. The marginal distribution of each coordinate remains unchanged, but the probability of compatible basin selection becomes
\begin{align}
    \Pr(b_k=b_{k+1})
    &=
    \Pr(\operatorname{sign}(\xi_k)=\operatorname{sign}(\xi_{k+1}))
    \notag \\
    &=
    \frac{1}{2}
    +
    \frac{1}{\pi}\arcsin(\chi).
\end{align}
Thus, any $\chi>0$ increases this probability over independent sampling. This abstraction does not aim to model the full nonlinear reverse process; instead, it isolates the coordination effect of correlated early randomness. It suggests that coupling the joint randomness across neighboring chunks can improve overlap compatibility while preserving the marginal noise distribution expected by each local denoiser.

Figure~\ref{fig:comparison_visualization_task1} shows a corresponding empirical instance in \texttt{AntMaze-Giant-Stitch}. Independently initialized chunks can enter incompatible overlap modes and become invalid after stitching, even when the individual chunks are locally plausible. This motivates MCDC: rather than changing the pretrained denoiser or stitching rule, we modify the joint initialization of chunk noises so that each chunk retains a standard Gaussian marginal while neighboring chunks are biased toward compatible overlap-mode choices. Additional mode-conflict visualizations are provided in Appendix~\ref{app:mode_conflict_visualizations}.

%% file: 5_Method.tex
\section{Method}

MCDC is a drop-in modification to the inference sampler of a pretrained compositional diffusion planner. The denoising network, reverse update, conditioning mechanism, and stitching rule are kept fixed. We first construct a coupled Gaussian initialization with exact per-chunk standard Gaussian marginals, and then insert it into the original compositional sampling pipeline.

\textbf{Notation.}
Let $k\in\{1,\ldots,K\}$ index trajectory chunks and $t\in\{0,\ldots,T\}$ index diffusion steps, with $t=T$ denoting the initial Gaussian sample and $t=0$ the final denoised sample. For $\tau_k^t$, the subscript indexes the chunk and the superscript indexes the diffusion step. We denote the base planner's independent initial noise by $\epsilon_k^{\mathrm{ind}}$ and the coupled initial noise constructed by MCDC by $\epsilon_k^{\mathrm{MCDC}}$.

\subsection{Marginal-Preserving Coupled Noise}

Let $\mathcal{G}_c=(\mathcal{V},\mathcal{E})$ denote the chunk adjacency graph induced by the overlapping chunk decomposition, where $\mathcal{V}=\{1,\ldots,K\}$ indexes trajectory chunks and $(i,j)\in\mathcal{E}$ if chunks $i$ and $j$ share an overlap region. In the sequential decomposition used in our experiments, $\mathcal{G}_c$ is a chain graph with $\mathcal{E}=\{(k,k+1)\}_{k=1}^{K-1}$. MCDC replaces the independent initial chunk noises $\{\epsilon_k^{\mathrm{ind}}\}_{k=1}^K$ with coupled noises $\{\epsilon_k^{\mathrm{MCDC}}\}_{k=1}^K$.

Let $\epsilon_{\mathrm{g}}$, $\{\epsilon_e\}_{e\in\mathcal{E}}$, and $\{\epsilon_{\mathrm{p},k}\}_{k=1}^K$ be mutually independent standard Gaussian noises with the same shape as a chunk. The global component $\epsilon_{\mathrm{g}}$ is shared across all chunks, each edge component $\epsilon_e$ is shared only by the two chunks incident to edge $e$, and each private component $\epsilon_{\mathrm{p},k}$ is used only by chunk $k$. For each chunk $k$, define
\begin{align}
    \epsilon_k^{\mathrm{MCDC}}
    =
    \sqrt{\rho_{\mathrm{g}}}\,\epsilon_{\mathrm{g}}
    &+
    \sum_{e\in\mathcal{E}: k\in e}
    \sqrt{\frac{\rho_{\mathrm{l}}}{d_k}}\,\epsilon_e \notag \\
    &+
    \sqrt{1-\rho_{\mathrm{g}}-\rho_{\mathrm{l}}\mathbf{1}_{\{d_k>0\}}}\,
    \epsilon_{\mathrm{p},k},
    \label{eq:mcdc_noise}
\end{align}
where $d_k$ is the degree of chunk $k$ in $\mathcal{G}_c$, and the coupling strengths satisfy
\begin{align}
    \rho_{\mathrm{g}}\geq 0,\qquad
    \rho_{\mathrm{l}}\geq 0,\qquad
    \rho_{\mathrm{g}}+\rho_{\mathrm{l}}\leq 1.
\end{align}
The normalization by $d_k$ assigns a total local-coupling variance of $\rho_{\mathrm{l}}$ to the incident edge components of each non-isolated chunk. Setting $\rho_{\mathrm{g}}=0$ gives an edge-local coupling variant, setting $\rho_{\mathrm{l}}=0$ gives a global coupling variant, and using both components gives the hybrid variant.

By construction, each $\epsilon_k^{\mathrm{MCDC}}$ preserves the same standard Gaussian marginal as $\epsilon_k^{\mathrm{ind}}$, while the shared components induce controlled cross-chunk covariance. Neighboring chunks share both the global component and their edge component, whereas non-neighboring chunks share only the global component when $\rho_{\mathrm{g}}>0$. The resulting marginal and covariance properties are formalized in Theorem~\ref{thm:mcdc_marginals}.

\subsection{MCDC Sampling for Compositional Planning}

MCDC is inserted only at the initial noise sampling step. Given the coupled noises in Eq.~\eqref{eq:mcdc_noise}, each chunk is initialized as
\begin{align}
    \tau_k^T = \epsilon_k^{\mathrm{MCDC}}.
\end{align}
The planner then follows the original compositional reverse update. Written in the standard stochastic notation,
\begin{align}
    p_\theta(\tau_k^{t-1}\mid \tau_k^t,c_k^t)
    =
    \mathcal{N}\!\left(
    \mu_\theta(\tau_k^t,t,c_k^t),
    \sigma_t^2 I
    \right),
\end{align}
where $c_k^t$ contains the start--goal, boundary, neighbor, or overlap conditions of the base planner. MCDC does not change $\epsilon_\theta$, $\mu_\theta$, $c_k^t$, or the final stitching rule.

In our implementation, MCDC is applied to compositional DDIM sampling by replacing only the initial chunk noises \cite{DDIM,CompDiffuser}. It introduces no learned verifier, additional guidance objective, or candidate search; the only extra computation is sampling the shared Gaussian components in Eq.~\eqref{eq:mcdc_noise}. Thus, MCDC is a marginal-preserving replacement for independent chunk initialization in pretrained compositional diffusion planners.

\subsection{Theoretical Analysis}

We formalize two properties of MCDC. The first shows that the coupled initialization preserves the standard Gaussian marginal expected by each pretrained local denoiser. The second explains, under a simplified overlap-basin abstraction, how positive cross-chunk correlation increases compatible mode selection. Because Eq.~\eqref{eq:mcdc_noise} constructs the coupled noises as a linear transformation of mutually independent Gaussian components, it defines a valid joint Gaussian distribution; equivalently, the induced block covariance matrix is positive semidefinite.

\begin{theorem}
\label{thm:mcdc_marginals}
For the coupled initial noise in Eq.~\eqref{eq:mcdc_noise},
\begin{align}
    \epsilon_k^{\mathrm{MCDC}} \sim \mathcal{N}(0,I).
\end{align}
Moreover, for any $i\neq j$,
\begin{align}
    \operatorname{Cov}(
    \epsilon_i^{\mathrm{MCDC}},
    \epsilon_j^{\mathrm{MCDC}})
    =
    \eta_{ij} I,
\end{align}
where
\begin{align}
    \eta_{ij}
    =
    \rho_{\mathrm{g}}
    +
    \mathbf{1}_{\{(i,j)\in\mathcal{E}\}}
    \frac{\rho_{\mathrm{l}}}{\sqrt{d_i d_j}} .
\end{align}
\end{theorem}

Theorem~\ref{thm:mcdc_marginals} gives the distributional basis of MCDC: each chunk starts from the same marginal distribution as in the base planner, while the joint distribution across chunks is no longer independent.

\begin{theorem}
\label{thm:mcdc_compatibility}
Consider an overlap edge $(i,j)\in\mathcal{E}$. Suppose the two neighboring chunks have two feasible overlap basins, and their basin choices are determined by scalar coordinates
\begin{align}
    \xi_i=\langle w,\epsilon_i^{\mathrm{MCDC}}\rangle,
    \qquad
    \xi_j=\langle w,\epsilon_j^{\mathrm{MCDC}}\rangle,
    \qquad
    \|w\|_2=1.
\end{align}
Then $(\xi_i,\xi_j)$ are jointly Gaussian with standard Gaussian marginals and correlation $\chi_{ij}=\eta_{ij}$. Let
\begin{align}
    M_i=\operatorname{sign}(\xi_i),
    \qquad
    M_j=\operatorname{sign}(\xi_j).
\end{align}
If compatibility requires the same overlap mode, then
\begin{align}
    \Pr(M_i=M_j)
    =
    \frac{1}{2}
    +
    \frac{1}{\pi}\arcsin(\chi_{ij}).
\end{align}
\end{theorem}

Thus, positive effective correlation increases the probability that adjacent chunks select compatible overlap basins. This captures the intended role of MCDC: coordinating early sampling directions across chunks without changing any local Gaussian marginal.

\begin{corollary}
\label{cor:mcdc_conflicts}
Under the binary overlap-basin model on every edge $(i,j)\in\mathcal{E}$, the expected number of overlap mode conflicts is
\begin{align}
    \operatorname{E}[\mathcal{C}_{\mathrm{conf}}(\boldsymbol{\tau})]
    =
    \sum_{(i,j)\in\mathcal{E}}
    \left(
    \frac{1}{2}
    -
    \frac{1}{\pi}\arcsin(\chi_{ij})
    \right).
\end{align}
\end{corollary}

The corollary lifts the pairwise mechanism to the trajectory level: increasing positive effective correlations on overlap edges reduces the expected number of local mode mismatches. Ordered multi-basin overlaps can be handled by replacing the binary sign partition with Gaussian quantile bins. The multi-basin extension and proofs are deferred to Appendix~\ref{app:proofs}.

%% file: 6_Experiments.tex
\begin{table*}[t]
	\centering
	\resizebox{0.95\textwidth}{!}{%
		\footnotesize
		\setlength{\tabcolsep}{4.5pt}
		\begin{tabular}{lll cccccccc}
		\toprule
		\textbf{Env} & \textbf{Type} & \textbf{Size} & \textbf{GCBC} & \textbf{GCIVL} & \textbf{GCIQL} & \textbf{HIQL} & \textbf{DD} & \textbf{CD} & \textbf{CDGS} & \textbf{MCDC} \\
		\midrule
		\multirow[c]{3}{*}{\texttt{Pointmaze}} & \multirow[c]{3}{*}{\texttt{Stitch}}
			& \texttt{Medium} & \valstd{23.2}{16.7} & \valstd{68.4}{12.4} & \valstd{22.6}{5.8} & \valstd{62.0}{6.8} & \valstd{76.0}{8.9} & \valstd{\mathbf{100.0}}{0.0} & \valstd{\mathbf{100.0}}{0.0} & \valstd{\mathbf{100.0}}{0.0} \\
			& & \texttt{Large} & \valstd{4.4}{8.3} & \valstd{8.0}{11.0} & \valstd{25.2}{3.6} & \valstd{22.0}{7.5} & \valstd{32.0}{11.0} & \valstd{99.4}{1.3} & \valstd{\mathbf{100.0}}{0.0} & \valstd{\mathbf{100.0}}{0.0} \\
			& & \texttt{Giant} & \valstd{0.0}{0.0} & \valstd{0.0}{0.0} & \valstd{0.0}{0.0} & \valstd{0.0}{0.0} & \valstd{0.0}{0.0} & \valstd{68.0}{3.2} & \valstd{71.6}{5.3} & \valstd{\mathbf{77.6}}{1.5} \\
		\midrule
		\multirow[c]{3}{*}{\texttt{Antmaze}} & \multirow[c]{3}{*}{\texttt{Stitch}}
			& \texttt{Medium} & \valstd{52.6}{5.7} & \valstd{32.6}{10.4} & \valstd{25.4}{6.5} & \valstd{89.8}{3.8} & \valstd{59.8}{3.3} & \valstd{94.0}{1.6} & \valstd{92.0}{1.6} & \valstd{\mathbf{98.0}}{1.0} \\
			& & \texttt{Large} & \valstd{1.8}{2.5} & \valstd{5.0}{4.9} & \valstd{7.4}{2.8} & \valstd{63.6}{8.6} & \valstd{0.2}{0.4} & \valstd{83.6}{2.5} & \valstd{85.2}{4.1} & \valstd{\mathbf{96.0}}{0.7} \\
			& & \texttt{Giant} & \valstd{0.0}{0.0} & \valstd{0.0}{0.0} & \valstd{0.0}{0.0} & \valstd{0.0}{0.0} & \valstd{0.0}{0.0} & \valstd{65.2}{1.9} & \valstd{\mathbf{78.4}}{3.4} & \valstd{76.6}{3.6} \\
		\midrule
		\multirow[c]{3}{*}{\texttt{Humanoidmaze}} & \multirow[c]{3}{*}{\texttt{Stitch}}
			& \texttt{Medium} & \valstd{26.2}{5.4} & \valstd{31.8}{6.1} & \valstd{12.0}{2.8} & \valstd{74.8}{3.0} & \valstd{63.4}{2.6} & \valstd{91.2}{2.5} & \valstd{89.4}{2.7} & \valstd{\mathbf{94.4}}{1.7} \\
			& & \texttt{Large} & \valstd{4.2}{1.7} & \valstd{5.0}{3.8} & \valstd{0.0}{0.0} & \valstd{23.6}{2.9} & \valstd{13.4}{5.1} & \valstd{73.8}{3.9} & \valstd{65.0}{4.7} & \valstd{\mathbf{79.8}}{3.3} \\
			& & \texttt{Giant} & \valstd{0.0}{0.0} & \valstd{0.0}{0.0} & \valstd{0.0}{0.0} & \valstd{0.0}{0.0} & \valstd{20.4}{2.3} & \valstd{42.0}{4.3} & \valstd{38.2}{2.6} & \valstd{\mathbf{54.6}}{2.6} \\
		\midrule
		\multirow[c]{2}{*}{\texttt{Antsoccer}} & \multirow[c]{2}{*}{\texttt{Stitch}}
			& \texttt{arena} & \valstd{21.6}{3.9} & \valstd{3.8}{1.8} & \valstd{13.2}{6.2} & \valstd{14.2}{2.8} & \valstd{20.4}{2.3} & \valstd{66.6}{2.4} & \valstd{60.8}{3.7} & \valstd{\mathbf{82.8}}{2.2} \\
			& & \texttt{medium} & \valstd{3.8}{2.8} & \valstd{0.4}{0.9} & \valstd{2.6}{1.1} & \valstd{4.0}{1.2} & \valstd{1.0}{1.0} & \valstd{9.4}{3.0} & \valstd{9.0}{3.4} & \valstd{\mathbf{15.0}}{1.2} \\
		\bottomrule
		\end{tabular}%
	}
    \caption{
    \textbf{Quantitative results on OGBench tasks.} We evaluate 5 goal-conditioned tasks per environment, with 20 episodes per task, and report mean $\pm$ standard deviation over 5 seeds.
    }
	\label{tab:main_result}
\end{table*}

\section{Experiments}

We evaluate MCDC on OGBench long-horizon goal-conditioned planning tasks under the compositional diffusion setting of CompDiffuser \cite{CompDiffuser}. The goal is to test whether replacing only the initial chunk noises improves planning while keeping the pretrained denoiser, reverse process, and stitching rule unchanged, without relying on search-based candidate selection. Further implementation details and ablations are provided in Appendices~\ref{app:impl_detail} and~\ref{app:additional_experiments}.

\paragraph{Datasets and environments.}
We evaluate on OGBench \texttt{Stitch} tasks that require composing short offline trajectory segments into long-horizon start--goal plans. The benchmark includes locomotion tasks with \texttt{PointMaze}, \texttt{AntMaze}, and \texttt{HumanoidMaze} across \texttt{Medium}, \texttt{Large}, and \texttt{Giant} maze sizes, and object-interaction tasks with \texttt{AntSoccer} in \texttt{Arena} and \texttt{Medium} settings. Since individual segments do not directly cover the test-time start--goal pairs, success depends on stitching locally feasible segments into a coherent long-horizon trajectory. Environment details and the evaluation protocol are provided in Appendix~\ref{app:environments}.

\paragraph{Baselines.}
We compare with offline GCRL baselines, including GCBC \cite{GCSL, WGCSL}, GCIVL, GCIQL \cite{IQL, OGBench}, and HIQL \cite{HIQL}, and diffusion-based planners, including Decision Diffuser (DD) \cite{DD}, CompDiffuser (CD) \cite{CompDiffuser}, and CDGS \cite{CDGS}. For CD, CDGS, and MCDC, we use the same pretrained local diffusion model for fair comparison.

\subsection{Main Results on OGBench}
\label{sec:main_results}

Table~\ref{tab:main_result} reports success rates on state-based OGBench \texttt{Stitch} tasks, including \texttt{Pointmaze}, \texttt{Antmaze}, \texttt{Humanoidmaze}, and \texttt{Antsoccer}. Standard offline GCRL methods are competitive on some medium-scale tasks, but degrade sharply on large and giant environments, where test-time start--goal pairs require composing behaviors beyond the length and coverage of individual offline trajectories. Diffusion-based planners are stronger in this regime because they generate and stitch short local trajectory segments.

Using the same pretrained local diffusion model as CompDiffuser, MCDC improves over CD across all non-saturated settings and achieves the best or tied-best performance on most tasks. The gains are especially clear on larger tasks, consistent with the increased risk of overlap-mode mismatch under independent chunk initialization. These results show that marginal-preserving noise coupling improves compositional diffusion planning at inference time without retraining, guidance optimization, or candidate search. Appendix~G.2 reports inference costs and results with wall-validity filtering.

\begin{figure*}[t]
    \centering
    \includegraphics[width=0.95\linewidth]{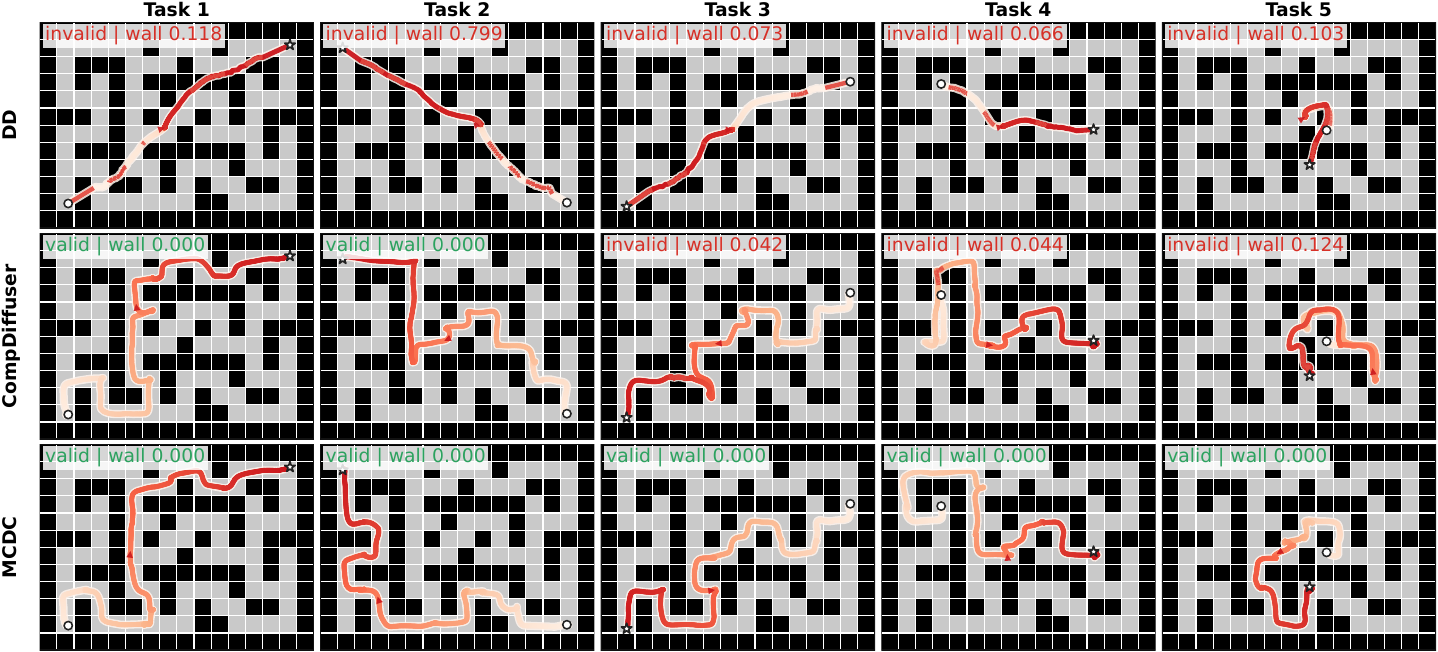}
    \caption{
    \textbf{Qualitative comparison on \texttt{AntMaze-Giant-Stitch}.} Each column shows one start--goal task, and each row shows the planned trajectory from DD, CompDiffuser, or MCDC. Green and red labels indicate wall-valid and wall-invalid plans, respectively.
    }
    \label{fig:antmaze_trajectory}
\end{figure*}

\subsection{Planned Trajectory Feasibility}
\label{sec:trajectory_feasibility}

We visualize planned trajectories before execution to assess whether composed plans respect maze geometry. As shown in Fig.~\ref{fig:antmaze_trajectory}, DD often generates direct but wall-invalid long-horizon plans, while CompDiffuser can still fail after stitching locally plausible chunks with incompatible overlap modes. MCDC yields wall-valid plans in the shown examples using the same pretrained local diffusion model, indicating improved compositional compatibility. Similar \texttt{PointMaze-Giant-Stitch} visualizations are provided in Appendix~\ref{app:planed_trajectory}. Appendix~\ref{app:first_plan_geometry} reports pre-execution wall-validity statistics for the original chunks and fused plans on both Giant mazes.

\subsection{Additional Results}
\label{sec:additional_results}

\paragraph{Coupling strategy ablation.}
Table~\ref{tab:coupling_ablation} compares global-only, local-only, and hybrid coupling. Local coupling is more reliable than global-only coupling on larger mazes, indicating that agreement along overlap edges is more useful than coarse global synchronization alone. Hybrid coupling further combines edge-local agreement with weak global coordination and achieves the strongest results in most non-saturated settings, while leaving the pretrained local denoiser unchanged.

\begin{table}[h]
    \centering
    \resizebox{0.49\textwidth}{!}{
    \small
    \setlength{\tabcolsep}{4pt}
    \begin{tabular}{lcccc}
    \toprule
    Dataset & CD & Global & Local & Hybrid \\
    \midrule
    \texttt{Point-M} & \valstd{\mathbf{100.0}}{0.0} & \valstd{\mathbf{100.0}}{0.0} & \valstd{\mathbf{100.0}}{0.0} & \valstd{\mathbf{100.0}}{0.0} \\
    \texttt{Point-L}  & \valstd{99.4}{1.3} & \valstd{94.4}{2.7} & \valstd{\mathbf{100.0}}{0.0} & \valstd{\mathbf{100.0}}{0.0} \\
    \texttt{Point-G}  & \valstd{68.0}{3.2} & \valstd{63.2}{3.1} & \valstd{73.2}{3.1} & \valstd{\mathbf{77.6}}{1.5} \\
    \midrule
    \texttt{Ant-M} & \valstd{94.0}{1.6} & \valstd{94.2}{1.5} & \valstd{93.2}{1.8} & \valstd{\mathbf{98.0}}{1.0} \\
    \texttt{Ant-L}  & \valstd{83.6}{2.5} & \valstd{85.8}{1.6} & \valstd{89.6}{3.6} & \valstd{\mathbf{96.0}}{0.7} \\
    \texttt{Ant-G}  & \valstd{65.2}{1.9} & \valstd{62.0}{3.2} & \valstd{72.4}{3.0} & \valstd{\mathbf{76.6}}{3.6} \\
    \bottomrule
    \end{tabular}
    }
    \caption{\textbf{Ablation of noise coupling strategies.} We report success rates on OGBench Stitch tasks using the same pretrained local diffusion model and inference protocol; Global and Local ablate one coupling component from the corresponding Hybrid setting.}
    \label{tab:coupling_ablation}
\end{table}

\begin{table}[h]
    \centering
    \resizebox{0.49\textwidth}{!}{
    \small
    \setlength{\tabcolsep}{4pt}
    \begin{tabular}{lcccc}
    \toprule
    Dataset & CD & MCDC & CDGS & CDGS w/ MCDC \\
    \midrule
    \texttt{Point-M} & \valstd{\mathbf{100.0}}{0.0} & \valstd{\mathbf{100.0}}{0.0} & \valstd{\mathbf{100.0}}{0.0} & \valstd{\mathbf{100.0}}{0.0} \\
    \texttt{Point-L}  & \valstd{99.4}{1.3} & \valstd{\mathbf{100.0}}{0.0} & \valstd{\mathbf{100.0}}{0.0} & \valstd{\mathbf{100.0}}{0.0} \\
    \texttt{Point-G}  & \valstd{68.0}{3.2} & \valstd{77.6}{1.5} & \valstd{71.6}{5.3} & \valstd{\mathbf{83.4}}{3.1} \\
    \midrule
    \texttt{Ant-M} & \valstd{94.0}{1.6} & \valstd{98.0}{1.0} & \valstd{92.0}{1.6} & \valstd{\mathbf{98.4}}{0.5} \\
    \texttt{Ant-L}  & \valstd{83.6}{2.5} & \valstd{\mathbf{96.0}}{0.7} & \valstd{85.2}{4.1} & \valstd{94.6}{1.7} \\
    \texttt{Ant-G}  & \valstd{65.2}{1.9} & \valstd{76.6}{3.6} & \valstd{78.4}{3.4} & \valstd{\mathbf{82.8}}{3.3} \\
    \bottomrule
    \end{tabular}
    }
    \caption{
    \textbf{Combining MCDC with CDGS.}
    We report success rates on OGBench Stitch tasks.
    CDGS w/ MCDC applies MCDC coupled initialization before CDGS guided search, using the same pretrained local diffusion model and without retraining.
    }
    \label{tab:mcdc_cdgs}
\end{table}

\paragraph{Compatibility with guided search.}
Table~\ref{tab:mcdc_cdgs} evaluates whether MCDC can be combined with inference-time guided search. Replacing CDGS's independent chunk initialization with MCDC improves CDGS on all non-saturated tasks, with clear gains on \texttt{PointMaze-Giant} and \texttt{AntMaze-Giant}. This indicates that marginal-preserving noise coupling is complementary to candidate-search methods rather than a substitute for a specific refinement procedure. Additional comparisons with related planning baselines are provided in Appendix~\ref{app:ecd_rcd_comparison}.

\begin{figure}[h]
    \centering
    \includegraphics[width=\columnwidth]{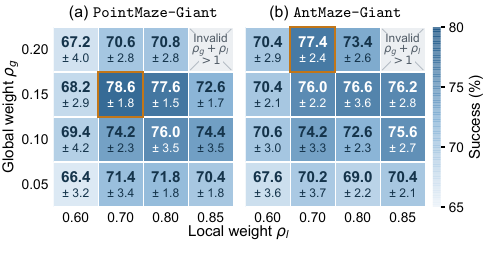}
    \vspace{-0.4cm}
    \caption{
    \textbf{Sensitivity to MCDC coupling strengths.}
    Cells report mean success rates (\%) $\pm$ standard deviations over five seeds. Columns vary $\rho_{\mathrm{l}}$ and rows vary $\rho_{\mathrm{g}}$. Orange outlines mark the highest observed mean in each panel.
    }
    \label{fig:rho_sensitivity}
\end{figure}

\paragraph{Correlation strength sensitivity.}
Fig.~\ref{fig:rho_sensitivity} shows the mean success rate of MCDC for 15 feasible combinations of local and global coupling strengths per environment. On both \texttt{PointMaze-Giant} and \texttt{AntMaze-Giant}, increasing $\rho_{\mathrm{l}}$ from 0.60 to 0.70 improved the mean success rate at every tested value of $\rho_{\mathrm{g}}$. The effect of further increases depended on both the global weight and the environment. For example, at $\rho_{\mathrm{g}}=0.15$, increasing $\rho_{\mathrm{l}}$ from 0.70 to 0.85 lowered the mean success rate on \texttt{PointMaze-Giant} from 78.6\% to 72.6\%. On \texttt{AntMaze-Giant}, the means across the same tested local weights ranged from 76.0\% to 76.6\%. The highest observed means occurred at $(\rho_{\mathrm{g}},\rho_{\mathrm{l}})=(0.15,0.70)$ and $(0.20,0.70)$, respectively. The shared setting used in our main experiments, $(0.15,0.80)$, was within 1.0 and 0.8 percentage points of the respective grid maxima.

%% file: 7_Conclusion.tex
\section{Conclusion}
We presented MCDC, a training-free inference method for compositional diffusion planning that replaces independent chunk-noise initialization with marginal-preserving coupled Gaussian noise. MCDC preserves the standard Gaussian marginal required by each pretrained local denoiser while introducing cross-chunk correlations that encourage compatible overlap-mode choices. Theoretical analysis under a multimodal basin abstraction and experiments on OGBench long-horizon stitch tasks show that this simple initialization-level modification improves compositional planning with negligible inference overhead. These results suggest that the joint initialization distribution is an effective and underexplored coordination channel for pretrained compositional diffusion planners, complementary to search- and guidance-based inference refinements.

%% file: 8_Appendix.tex
\section{Limitations and Future Work}
\label{app:limitations}

MCDC is a training-free modification of the sampling procedure, and therefore its effectiveness is bounded by the quality of the pretrained local diffusion model. If the local denoiser cannot generate feasible chunks for a given region, or if the required transition is not covered by the offline data, coupling the initial noises alone cannot recover missing local capabilities. MCDC also keeps the original conditioning and stitching rules unchanged, so failures caused by inaccurate boundary conditions, poor inverse dynamics execution, or suboptimal stitching may still remain.

Another limitation is that the coupling strengths are currently selected as fixed inference-time hyperparameters. Although this keeps the method simple and does not require additional training, different environments or task lengths may benefit from different levels of global and local coupling. Future work could adapt the coupling structure or strength based on the chunk graph, uncertainty of the local denoiser, or online diagnostics of overlap consistency. Extending MCDC beyond chain-like trajectory chunks to more general compositional structures is also a promising direction.

\section{Corridor-Basin Toy Setting}
\label{app:toy_setting}

The toy example in Fig.~\ref{fig:toy} is a controlled diagnostic for overlap-mode conflict in compositional diffusion sampling. It uses a synthetic corridor-basin setting with two feasible overlap modes, denoted as the upper and lower basins, separated by an invalid region. Each local chunk can select either basin, and both choices are locally feasible. A valid composition, however, requires neighboring chunks to select the same basin in their shared overlap. If adjacent chunks select different basins, the stitched trajectory crosses the invalid region or creates a low-density transition, although each local chunk is individually plausible.

The toy abstracts early denoising by scalar Gaussian mode coordinates. For each chunk $k$, the overlap mode is assigned by the sign of its initial coordinate, with positive and negative signs corresponding to the upper and lower basins. Independent sampling draws chunk noises independently, whereas MCDC couples neighboring chunk noises through shared Gaussian components while preserving each chunk's standard Gaussian marginal. Thus, the toy isolates whether changing cross-chunk dependence can improve compatibility without changing local marginals.

\begin{table}[h]
    \centering
    \small
    \begin{tabular}{ll}
    \toprule
    \textbf{Item} & \textbf{Setting} \\
    \midrule
    Number of chunks & $K=6$ \\
    Number of sampled compositions & $20000$ \\
    Random seed & $0$ \\
    Independent initialization & $\rho_{\mathrm{g}}=0,\ \rho_{\mathrm{l}}=0$ \\
    Hybrid MCDC & $\rho_{\mathrm{g}}=0.15,\ \rho_{\mathrm{l}}=0.7$ \\
    Local marginal metric & fraction of chunks assigned to the upper basin \\
    Pair compatibility metric & fraction of adjacent chunk pairs selecting the same basin \\
    Full-chain validity metric & fraction of samples with no adjacent basin mismatch \\
    \bottomrule
    \end{tabular}
    \caption{\textbf{Configuration of the corridor-basin toy.} The toy uses the same notation as the main method, where $\rho_{\mathrm{g}}$ and $\rho_{\mathrm{l}}$ denote global and local coupling strengths.}
    \label{tab:toy_config}
\end{table}

We report five diagnostics: the local basin marginal, the joint basin distribution, the probability that neighboring chunks select the same overlap mode, the fraction of fully valid stitched trajectories, and the compatibility probability as a function of the effective noise correlation. These diagnostics test whether coupling improves cross-chunk compatibility while preserving the local basin marginal. The toy is used only as a controlled diagnostic of the mode-conflict mechanism; the main empirical evaluation is conducted on OGBench tasks in Section~\ref{sec:main_results}.

\section{Theorem and Corollary Proofs}
\label{app:proofs}
\subsection{Proof of Theorem~\ref{thm:mcdc_marginals}}
\begin{proof}
The coupled noises are linear combinations of mutually independent Gaussian tensors. Therefore the joint vector
$(\epsilon_1^{\mathrm{MCDC}},\ldots,\epsilon_K^{\mathrm{MCDC}})$
is Gaussian. Its block covariance matrix is positive semidefinite because, for any coefficients $\{a_k\}_{k=1}^K$,
\[
\operatorname{Var}\left(\sum_{k=1}^K a_k^\top \epsilon_k^{\mathrm{MCDC}}\right)\ge 0.
\]
Thus Eq.~\eqref{eq:mcdc_noise} defines a valid joint Gaussian distribution.

Since $\epsilon_{\mathrm{g}}$, $\{\epsilon_e\}_{e\in\mathcal{E}}$, and
$\{\epsilon_{\mathrm{p},k}\}_{k=1}^K$ are mutually independent standard
Gaussian tensors, $\epsilon_k^{\mathrm{MCDC}}$ is Gaussian with zero mean.
By independence, its covariance is
\begin{align}
    \operatorname{Var}(\epsilon_k^{\mathrm{MCDC}})
    &=
    \rho_{\mathrm{g}} I
    +
    \sum_{e\in\mathcal{E}:k\in e}
    \frac{\rho_{\mathrm{l}}}{d_k} I
    +
    \left(
    1-\rho_{\mathrm{g}}
    -\rho_{\mathrm{l}}\mathbf{1}_{\{d_k>0\}}
    \right) I
    \notag \\
    &=
    \rho_{\mathrm{g}} I
    +
    \rho_{\mathrm{l}}\mathbf{1}_{\{d_k>0\}} I
    +
    \left(
    1-\rho_{\mathrm{g}}
    -\rho_{\mathrm{l}}\mathbf{1}_{\{d_k>0\}}
    \right) I
    \notag \\
    &= I.
\end{align}
Thus $\epsilon_k^{\mathrm{MCDC}}\sim\mathcal{N}(0,I)$.

For $i\neq j$, all private noises are independent, and distinct edge noises
are independent. Therefore the only shared components between
$\epsilon_i^{\mathrm{MCDC}}$ and $\epsilon_j^{\mathrm{MCDC}}$ are the
global noise $\epsilon_{\mathrm{g}}$ and, when $(i,j)\in\mathcal{E}$, the
edge noise $\epsilon_{(i,j)}$. Hence
\begin{align}
    \operatorname{Cov}(
    \epsilon_i^{\mathrm{MCDC}},
    \epsilon_j^{\mathrm{MCDC}})
    &=
    \rho_{\mathrm{g}} I
    +
    \mathbf{1}_{\{(i,j)\in\mathcal{E}\}}
    \frac{\rho_{\mathrm{l}}}{\sqrt{d_i d_j}} I
    \notag \\
    &=
    \eta_{ij} I.
\end{align}
\end{proof}

\subsection{Proof of Theorem~\ref{thm:mcdc_compatibility}}

\begin{proof}
For zero-mean jointly Gaussian variables $(\xi_i,\xi_j)$ with standard
Gaussian marginals and correlation $\chi_{ij}$, the Gaussian sign identity gives
\begin{align}
    \operatorname{E}\!\left[
    \operatorname{sign}(\xi_i)
    \operatorname{sign}(\xi_j)
    \right]
    =
    \frac{2}{\pi}\arcsin(\chi_{ij}).
\end{align}
By definition, $M_i=\operatorname{sign}(\xi_i)$ and
$M_j=\operatorname{sign}(\xi_j)$. Hence
$\operatorname{sign}(\xi_i)\operatorname{sign}(\xi_j)=1$ when
$M_i=M_j$, and equals $-1$ otherwise. Therefore,
\begin{align}
    \operatorname{E}\!\left[
    \operatorname{sign}(\xi_i)
    \operatorname{sign}(\xi_j)
    \right]
    =
    2\Pr(M_i=M_j)-1.
\end{align}
Rearranging gives
\begin{align}
    \Pr(M_i=M_j)
    =
    \frac{1}{2}
    +
    \frac{1}{\pi}\arcsin(\chi_{ij}).
\end{align}
\end{proof}

\subsection{Proof of Corollary~\ref{cor:mcdc_conflicts}}

\begin{proof}
For each overlap edge $(i,j)\in\mathcal{E}$, Theorem~\ref{thm:mcdc_compatibility}
gives
\begin{align}
    \Pr(M_i\neq M_j)
    =
    \frac{1}{2}
    -
    \frac{1}{\pi}\arcsin(\chi_{ij}).
\end{align}
The number of conflicting overlap edges can be written as
\begin{align}
    \mathcal{C}_{\mathrm{conf}}(\boldsymbol{\tau})
    =
    \sum_{(i,j)\in\mathcal{E}}
    \mathbf{1}_{\{M_i\neq M_j\}} .
\end{align}
Taking expectations and using linearity of expectation,
\begin{align}
    \operatorname{E}[\mathcal{C}_{\mathrm{conf}}(\boldsymbol{\tau})]
    &=
    \sum_{(i,j)\in\mathcal{E}}
    \operatorname{E}\!\left[
    \mathbf{1}_{\{M_i\neq M_j\}}
    \right]
    \notag \\
    &=
    \sum_{(i,j)\in\mathcal{E}}
    \Pr(M_i\neq M_j)
    \notag \\
    &=
    \sum_{(i,j)\in\mathcal{E}}
    \left(
    \frac{1}{2}
    -
    \frac{1}{\pi}\arcsin(\chi_{ij})
    \right).
\end{align}
This gives the stated result.
\end{proof}

\subsection{Ordered Multi-Basin Extension}
\label{app:multimode_theory}

\begin{corollary}[Multi-Basin Compatibility]
\label{cor:multibasin_compatibility}
Let $(\xi_i,\xi_j)$ be jointly Gaussian scalar basin coordinates with
standard Gaussian marginals and correlation $\chi_{ij}\in[0,1)$. For
$N\geq 2$, partition the real line into $N$ equal-probability ordered
Gaussian bins
\begin{align}
    \mathcal{B}_{\ell}=(\nu_{\ell-1},\nu_{\ell}],
    \qquad
    \ell=1,\ldots,N,
\end{align}
where
\begin{align}
    \nu_0=-\infty,\qquad
    \nu_N=+\infty,\qquad \notag \\
    \nu_{\ell}=\Phi^{-1}\left(\frac{\ell}{N}\right),
    \quad \ell=1,\ldots,N-1,
\end{align}
and $\Phi$ is the standard Gaussian CDF. Define $M_i=\ell$ if
$\xi_i\in\mathcal{B}_{\ell}$ and $M_j=\ell$ if
$\xi_j\in\mathcal{B}_{\ell}$. Then
\begin{align}
    p_{\mathrm{same}}^{(N)}(\chi_{ij})
    &=
    \Pr(M_i=M_j)
    \notag \\
    &=
    \sum_{\ell=1}^N
    \Pr(\xi_i\in\mathcal{B}_{\ell},
    \xi_j\in\mathcal{B}_{\ell}).
\end{align}
Moreover,
\begin{align}
    p_{\mathrm{same}}^{(N)}(0)=\frac{1}{N},
    \qquad
    \lim_{\chi_{ij}\to 1}
    p_{\mathrm{same}}^{(N)}(\chi_{ij})=1,
\end{align}
and $p_{\mathrm{same}}^{(N)}(\chi_{ij})$ is nondecreasing for
$\chi_{ij}\in[0,1)$.
\end{corollary}

\begin{proof}
The expression for $p_{\mathrm{same}}^{(N)}$ follows by summing the
bivariate Gaussian probability over the diagonal rectangles
$\mathcal{B}_{\ell}\times\mathcal{B}_{\ell}$. When $\chi_{ij}=0$, the two
scalar coordinates are independent and each bin has probability $1/N$.
Therefore,
\begin{align}
    p_{\mathrm{same}}^{(N)}(0)
    =
    \sum_{\ell=1}^N
    \left(\frac{1}{N}\right)^2
    =
    \frac{1}{N}.
\end{align}

It remains to show monotonicity. By Mehler's expansion, for
$\chi_{ij}\in[0,1)$,
\begin{align}
    \phi_2(u,v;\chi_{ij})
    =
    \phi(u)\phi(v)
    \sum_{m=0}^{\infty}
    \frac{\chi_{ij}^m}{m!}
    H_m(u)H_m(v),
\end{align}
where $\phi_2$ is the bivariate standard Gaussian density with correlation
$\chi_{ij}$, $\phi$ is the univariate standard Gaussian density, and
$H_m$ denotes the probabilists' Hermite polynomial. Hence
\begin{align}
    p_{\mathrm{same}}^{(N)}(\chi_{ij})
    =
    \sum_{\ell=1}^{N}
    \sum_{m=0}^{\infty}
    \frac{\chi_{ij}^m}{m!}
    \left(
    \int_{\mathcal{B}_{\ell}}
    H_m(u)\phi(u)\,du
    \right)^2 .
\end{align}
All coefficients in this power series are nonnegative, so
$p_{\mathrm{same}}^{(N)}(\chi_{ij})$ is nondecreasing on
$\chi_{ij}\in[0,1)$.

Finally, as $\chi_{ij}\to 1$, the jointly Gaussian pair satisfies
$\xi_j\to\xi_i$ in probability. Since Gaussian bin boundaries have
probability zero, the probability that $\xi_i$ and $\xi_j$ fall into
different bins converges to zero. Therefore,
\begin{align}
    \lim_{\chi_{ij}\to 1}
    p_{\mathrm{same}}^{(N)}(\chi_{ij})=1.
\end{align}
\end{proof}

\section{Practical Algorithm}
\label{app:algorithm}
Algorithm~\ref{alg:mcdc} summarizes the sampling procedure. MCDC only replaces the independent initial noises with the coupled noises in Eq.~\eqref{eq:mcdc_noise}; the pretrained denoiser and stitching procedure are unchanged \cite{CompDiffuser}.

\begin{algorithm}[h]
\caption{MCDC Sampling}
\label{alg:mcdc}
\begin{algorithmic}[1]
\REQUIRE Pretrained denoising network $\epsilon_\theta$, chunk graph $\mathcal{G}_c=(\mathcal{V},\mathcal{E})$, condition $(s_0,g)$, coupling strengths $\rho_{\mathrm{g}},\rho_{\mathrm{l}}$
\ENSURE Composed trajectory $\boldsymbol{\tau}$

\STATE Sample coupled initial noises $\{\epsilon_k^{\mathrm{MCDC}}\}_{k=1}^K$ by Eq.~\eqref{eq:mcdc_noise}
\STATE Initialize each chunk by $\tau_k^T \leftarrow \epsilon_k^{\mathrm{MCDC}}$

\FOR{$t=T,T-1,\ldots,1$}
    \FOR{$k=1,\ldots,K$}
        \STATE Construct local condition $c_k^t$ from boundary and overlap information
        \STATE Compute denoising mean $\mu_\theta(\tau_k^t,t,c_k^t)$ from $\epsilon_\theta(\tau_k^t,t,c_k^t)$
        \STATE Sample $\tau_k^{t-1}\sim \mathcal{N}(\mu_\theta(\tau_k^t,t,c_k^t),\sigma_t^2I)$
    \ENDFOR
\ENDFOR

\STATE Stitch $\{\tau_k^0\}_{k=1}^K$ through overlap regions to obtain $\boldsymbol{\tau}$
\RETURN $\boldsymbol{\tau}$
\end{algorithmic}
\end{algorithm}

\section{Environments and Datasets}
\label{app:environments}
\begin{figure}[h]
    \centering
    \includegraphics[width=0.48\linewidth]{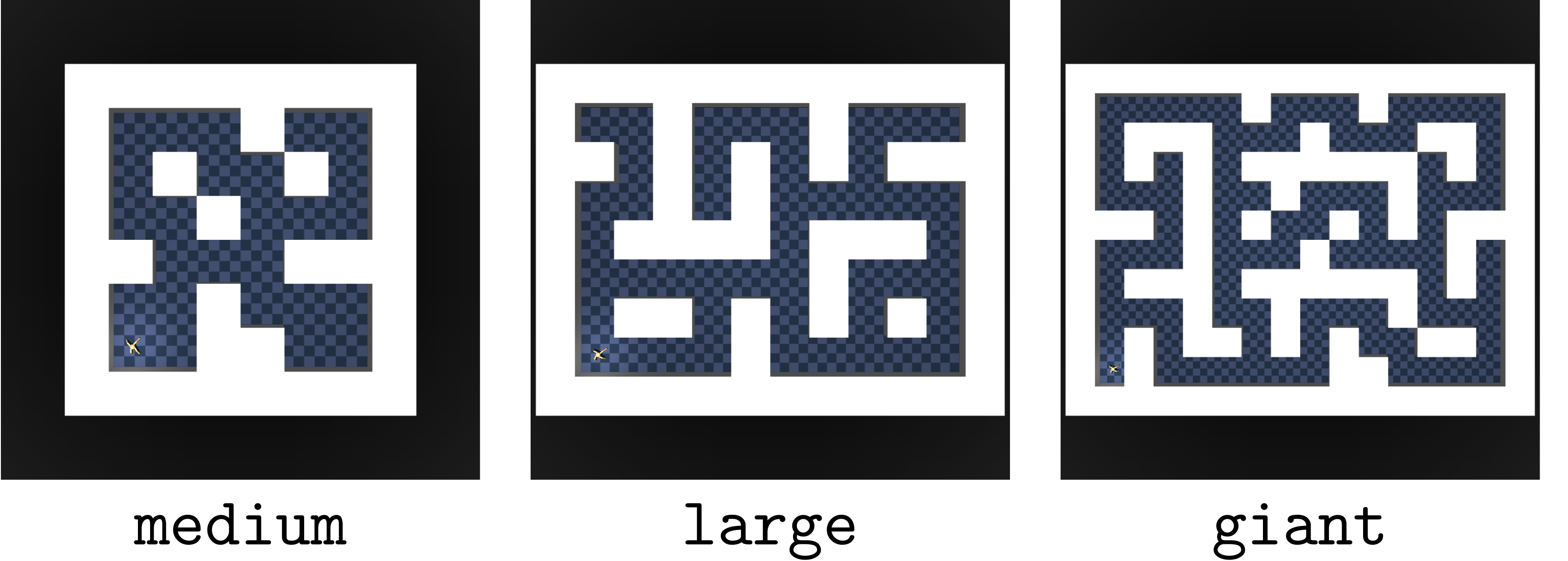}
    \includegraphics[width=0.48\linewidth]{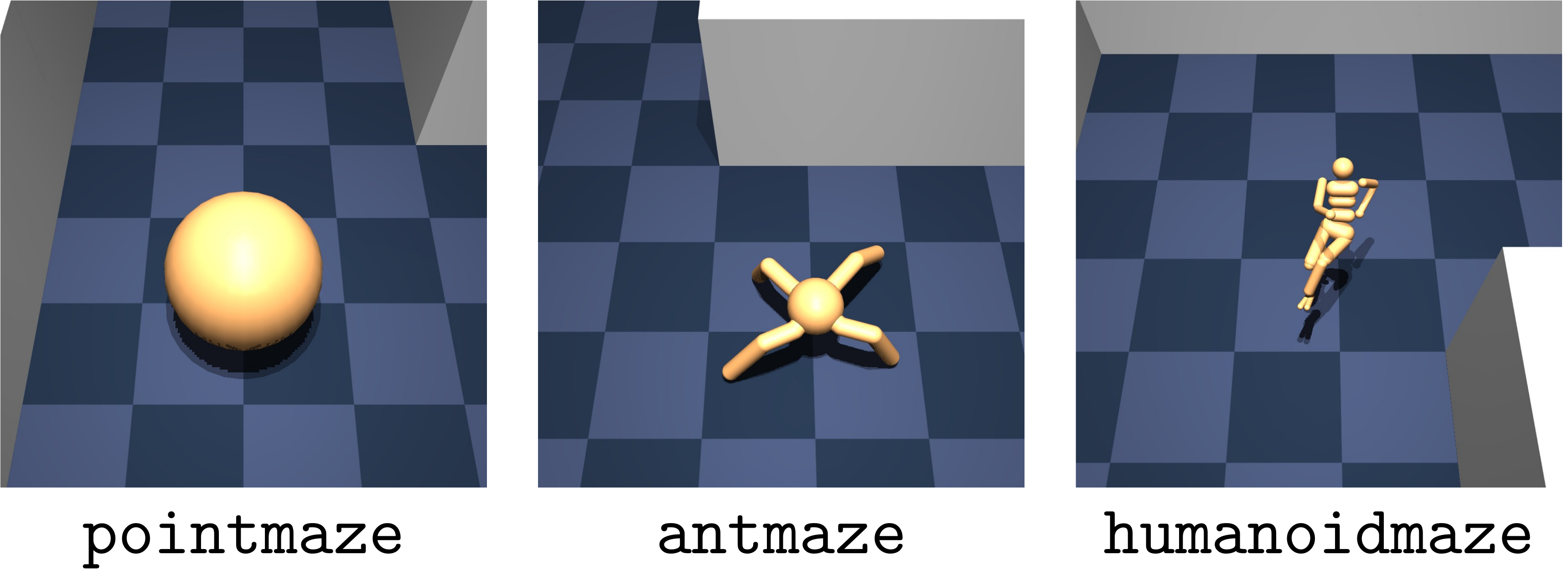}
    \includegraphics[width=0.48\linewidth]{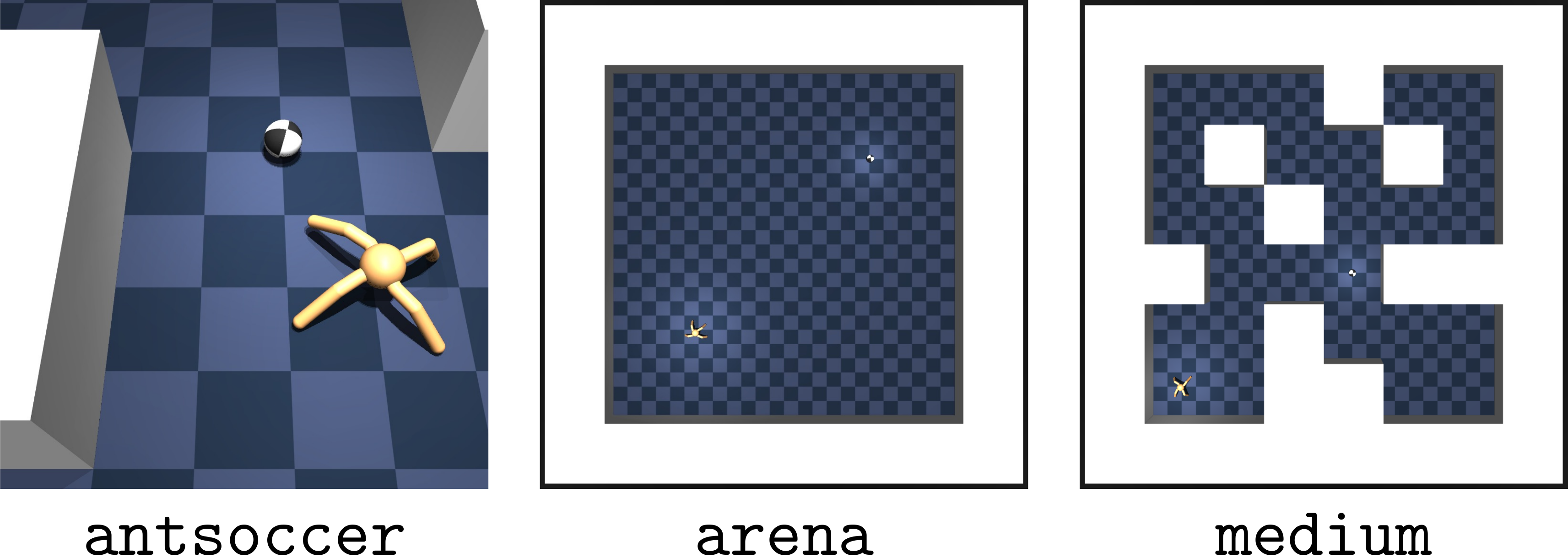}
    \caption{Illustration of the OGBench environments used in our evaluation, covering locomotion mazes and AntSoccer tasks.}
	\label{fig:environments}
\end{figure}
\paragraph{Maze.}
The maze tasks require an agent to navigate from a given start state to a specified goal state under sparse rewards. We consider three agent types with increasing dynamics complexity: \texttt{pointmaze}, \texttt{antmaze}, and \texttt{humanoidmaze}. The corresponding agents are a 2-DoF point mass, an 8-DoF quadruped Ant, and a 21-DoF Humanoid, respectively. For each agent type, we evaluate \texttt{medium}, \texttt{large}, and \texttt{giant} Stitch datasets. These datasets consist of short trajectory fragments that do not necessarily connect test-time start-goal pairs, making long-horizon success dependent on stitching locally feasible segments into a globally valid trajectory.

\paragraph{Antsoccer.}
The \texttt{antsoccer} tasks require an Ant robot to move a ball to a target location. Compared with \texttt{antmaze}, these tasks combine navigation with object interaction, making the composed trajectory more sensitive to local planning errors. We evaluate two Stitch variants: \texttt{arena}, an open layout, and \texttt{medium}, a maze-like layout with obstacles. These tasks further test whether a planner can compose short-horizon behaviors into long-horizon goal-reaching trajectories.

\begin{table}[h]
	\centering
	\setlength{\tabcolsep}{8pt}
	\resizebox{0.48\textwidth}{!}{
    \begin{tabular}{lllc}
		\toprule
		\textbf{Environment}                   & \textbf{Type}                       & \textbf{Size}                & \textbf{Env Steps} \\
		\midrule
		\multirow[c]{3}{*}{\texttt{Pointmaze}} & \multirow[c]{3}{*}{\texttt{Stitch}}
		                                                                              & \texttt{Medium}               & 1000                                              \\
		                                       &                                     & \texttt{Large}                & 1000               \\
		                                       &                                     & \texttt{Giant}                & 1000               \\
		\midrule
		\multirow[c]{3}{*}{\texttt{Antmaze}}   & \multirow[c]{3}{*}{\texttt{Stitch}}
		                                                                            & \texttt{Medium}               & 1000                                              \\
		                                       &                                     & \texttt{Large}                & 1000               \\
		                                       &                                     & \texttt{Giant}                & 2000               \\
		\midrule
		\multirow[c]{3}{*}{\makecell[l]{\texttt{Humanoid}\\\texttt{maze}}} & \multirow[c]{3}{*}{\texttt{Stitch}}
		                                                                              & \texttt{Medium}               & 5000                                              \\
		                                       &                                     & \texttt{Large}                & 5000               \\
		                                       &                                     & \texttt{Giant}                & 8000               \\
		\midrule
		\multirow[c]{2}{*}{\texttt{Antsoccer}} & \multirow[c]{2}{*}{\texttt{Stitch}}
		                                                                              & \texttt{Arena}                & 5000                                              \\
		                                       &                                     & \texttt{Medium}               & 5000               \\ 
        \bottomrule
	\end{tabular}
    }
    \caption{\textbf{Evaluation environments and datasets in OGBench.} We follow the evaluation setup of CompDiffuser~\cite{CompDiffuser}.}
	\label{tab:env_datasets}
\end{table}

Table~\ref{tab:env_datasets} lists all evaluation environments with their corresponding OGBench dataset names and the maximum environment steps per episode.

\section{Implementation Details}
\label{app:impl_detail}

\subsection{Diffusion Model Architecture and Training}
\label{app:arch}
Our implementation builds on the public CompDiffuser codebase\footnote{\url{https://github.com/devinluo27/comp_diffuser_release}}~\cite{CompDiffuser}. 
For fair comparison, MCDC uses the same pretrained local diffusion models and evaluation protocol as CompDiffuser; no model parameters are updated when applying MCDC.
For \texttt{pointmaze}, \texttt{antmaze}, and \texttt{humanoidmaze}, the local denoiser is a 1D temporal U-Net with residual blocks, group normalization, and sinusoidal timestep embeddings.
For \texttt{antsoccer}, we use the DiT-based architecture adopted by CompDiffuser, with environment-specific hidden dimensions and patch sizes.
The local diffusion models are trained with the standard DDPM objective and a linear noise schedule using Adam with learning rate $2\times 10^{-4}$.

\subsection{Evaluation Protocol}
\label{app:eval_protocol}
We follow the standard OGBench evaluation protocol~\cite{OGBench}. Each environment provides 5 test-time start-goal pairs, and we run 20 episodes for each pair. We report the binary success rate, where an episode is successful if the agent reaches within the environment-specific threshold distance of the goal. All results are averaged over 5 random seeds and reported as mean $\pm$ standard deviation. For a fair comparison under the same implementation and evaluation setup, we rerun all baselines.

For experiments with replanning, we follow the adaptive replanning strategy used in CompDiffuser~\cite{CompDiffuser}. Replanning is triggered when the agent deviates from the current subgoal beyond a threshold distance. The planner then applies a receding composition scheme that reduces the remaining number of chunks according to the progress already made along the current plan.

\subsection{Hyperparameters}
\label{app:mcdc_hyperparams}
Unless otherwise specified, all experiments use the hyperparameters summarized in Table~\ref{tab:impl_hyperparams}. 
We keep the local diffusion architecture, segment length, overlap size, and diffusion steps consistent with the CompDiffuser setup, so that the comparison isolates the effect of MCDC's coupled initialization. 
The only MCDC-specific hyperparameters are the local and global coupling strengths, $\rho_{\mathrm{l}}$ and $\rho_{\mathrm{g}}$, which are fixed for each environment family and are not tuned per task, seed, or start-goal pair. 
All selected values satisfy the validity condition $\rho_{\mathrm{l}}+\rho_{\mathrm{g}}\leq 1$. 
The sensitivity analysis in Fig.~\ref{fig:rho_sensitivity} performs local sweeps around these selected settings rather than exhaustive hyperparameter search.

\begin{table}[h]
	\centering
	\resizebox{0.68\textwidth}{!}{%
		\begin{tabular}{@{}l cccc}
			\toprule
			\textbf{Parameter}                     & \textbf{PointMaze} & \textbf{AntMaze} & \textbf{HumanoidMaze} & \textbf{AntSoccer}\\
			\midrule
			Architecture                           & U-Net & U-Net & U-Net & DiT  \\
			Segment length $H$                     & 160 & 160 & 336 & 160   \\
			Overlap size                           & 56 & 56 & 128 & 56   \\
			Diffusion steps $T$                    & 512 & 512 & 512 & 512  \\
            \midrule
            $\rho_{\mathrm{l}}$                    & 0.8 & 0.8 & 0.8 & 0.7  \\
            $\rho_{\mathrm{g}}$                    & 0.15 & 0.15 & 0.15 & 0.1  \\
			\bottomrule
		\end{tabular}%
	}
    \caption{\textbf{Hyperparameters across environments.}}
	\label{tab:impl_hyperparams}
\end{table}

Table~\ref{tab:env_n_segments} lists the number of composed trajectory segments used in each environment. 
These values follow the CompDiffuser evaluation setup and define the chunk adjacency graph used by MCDC.

\begin{table}[ht]
	\centering
	\resizebox{0.49\textwidth}{!}{%
    	\begin{tabular}{lllc}
    		\toprule
    		\textbf{Environment}                   & \textbf{Type}                       & \textbf{Size}                 & \textbf{\# Segments $K$} \\
    		\midrule
    		\multirow[c]{3}{*}{\texttt{Pointmaze}} & \multirow[c]{3}{*}{\texttt{Stitch}}
    		                                       & \texttt{Medium}              & 3                        \\
    		                                       &                                     & \texttt{Large}               & 6                        \\
    		                                       &                                     & \texttt{Giant}               & 8                        \\
    		\midrule
    		\multirow[c]{3}{*}{\texttt{Antmaze}}   & \multirow[c]{3}{*}{\texttt{Stitch}}
    		                                       & \texttt{Medium}              & 3                        \\
    		                                       &                                     & \texttt{Large}               & 6                        \\
    		                                       &                                     & \texttt{Giant}               & 9                        \\
    		\midrule
    		\multirow[c]{3}{*}{\makecell[l]{\texttt{Humanoid}                                                                                                                      \\\texttt{maze}}} & \multirow[c]{3}{*}{\texttt{Stitch}}
    		                                       & \texttt{Medium}              & 4                        \\
    		                                       &                                     & \texttt{Large}               & 6                        \\
    		                                       &                                     & \texttt{Giant}               & 11                       \\
    		\midrule
    		\multirow[c]{2}{*}{\texttt{Antsoccer}} & \multirow[c]{2}{*}{\texttt{Stitch}}
    		                                       & \texttt{Arena}               & 5                        \\
    		                                       &                                     & \texttt{Medium}              & 6                        \\
    		\bottomrule
    	\end{tabular}%
	}
    \caption{\textbf{Number of composed segments $K$ per environment.} For locomotion mazes and \texttt{antsoccer}, values follow the same setting as CompDiffuser~\cite{CompDiffuser}.}
	\label{tab:env_n_segments}
\end{table}

\subsection{Computational Resources}
\label{app:compute}
All experiments were conducted on a cluster equipped with four NVIDIA GeForce RTX 4090 GPUs, each with 24GB of VRAM. Unless otherwise specified, all reported inference times correspond to runs executed on a single GPU.

\subsection{Inverse Dynamic as Low-level Controllers}
\label{app:low_level}
The diffusion planner generates state-space trajectories, which are converted into executable actions by a separate low-level controller. 
Following CompDiffuser~\cite{CompDiffuser}, we use an MLP-based inverse dynamics model for locomotion and \texttt{antsoccer} tasks. 
The model takes consecutive states $(s_t, s_{t+1})$ as input and predicts the corresponding action $a_t$. 
We use the same architecture and training procedure as CompDiffuser for fair comparison.

\section{Additional Quantitative Experiments}
\label{app:additional_experiments}

\subsection{Ablation experiments on replanning}
\label{app:ablation_replan}
\begin{table}[h]
    \centering
	\resizebox{0.6\textwidth}{!}{%
		\begin{tabular}{ll cc cc}
			\toprule
			\multirow{2}{*}[-0.7ex]{\textbf{Env}} & \multirow{2}{*}[-0.7ex]{\textbf{Size}} & \multicolumn{2}{c}{\textbf{CD}} & \multicolumn{2}{c}{\textbf{MCDC}}                                              \\
			\cmidrule(lr){3-4} \cmidrule(lr){5-6}
			                                      &                                        & \textbf{\xmark}                 & \textbf{\cmark}                  & \textbf{\xmark} & \textbf{\cmark}          \\
			\midrule
			\multirow[c]{2}{*}{\texttt{pointmaze}}
			                                      & \texttt{Large}                         & \valstd{98.2}{1.8}                 & \valstd{99.4}{1.3}                   & \valstd{\mathbf{100.0}}{0.0}  & \valstd{\mathbf{100.0}}{0.0} \\
			                                      & \texttt{Giant}                         & \valstd{52.6}{4.6}                  & \valstd{68.0}{3.2}                   & \valstd{60.6}{4.9}  & \valstd{\mathbf{77.6}}{1.5} \\
			\midrule
			\multirow[c]{2}{*}{\texttt{antmaze}}
			                                      & \texttt{Large}                         & \valstd{69.8}{5.7}                  & \valstd{83.6}{2.5}                   & \valstd{75.0}{4.1}  & \valstd{\mathbf{96.0}}{0.7}  \\
			                                      & \texttt{Giant}                         & \valstd{28.4}{4.4}                  & \valstd{65.2}{1.9}                   & \valstd{42.8}{3.7}  & \valstd{\mathbf{76.6}}{3.6}  \\
			\midrule
			\multicolumn{2}{c}{\textbf{Average}}  & $62.3$                                 & $79.1$                          & $69.6$                           & $\mathbf{87.6}$                            \\
			\bottomrule
		\end{tabular}%
	}
    \caption{\textbf{Ablation study on replanning.} Success rates on \texttt{pointmaze} and \texttt{antmaze} \texttt{stitch}, comparing CD and MCDC with (\cmark) and without (\xmark) replanning.}
	\label{tab:ablation_replan}
\end{table}

Replanning regenerates a plan from the current state during execution to mitigate drift from the planned trajectory; details are provided in Appendix~\ref{app:eval_protocol}. Table~\ref{tab:ablation_replan} compares CD and MCDC with and without replanning on representative \texttt{PointMaze-Stitch} and \texttt{AntMaze-Stitch} tasks. Replanning improves both methods, with the largest gains on \texttt{Giant} mazes where long plans are more susceptible to execution drift.

MCDC consistently improves over CD under both settings. The advantage remains visible even without replanning, indicating that MCDC improves the quality of the composed plan itself rather than merely relying on online recovery during execution. With replanning enabled, MCDC further benefits from the same execution-time correction mechanism and achieves the strongest overall performance.

\subsection{Inference-time comparison}
\label{app:inference_time}
\begin{table}[h]
	\centering
	\resizebox{0.55\textwidth}{!}{%
		\begin{tabular}{l ll ll}
			\toprule
			\multirow{2}{*}[-0.7ex]{\textbf{Method}} & \multicolumn{2}{c}{\texttt{Pointmaze-Giant}} & \multicolumn{2}{c}{\texttt{Antmaze-Giant}}                                                        \\
			\cmidrule(lr){2-3} \cmidrule(lr){4-5}
			& \textbf{Success (\%)}                               & \textbf{Time (s)}                          & \textbf{Success (\%)}          & \textbf{Time (s)}          \\
			\midrule
			CD & \valstd{68.0}{3.2} & \valstd{10.4}{0.7} & \valstd{65.2}{1.9} & \valstd{11.6}{0.9}         \\
			CDGS & \valstd{71.6}{5.3} & \valstd{58.1}{5.1} & \valstd{\mathbf{78.4}}{3.4}  & \valstd{73.7}{9.6}     \\
			MCDC & \valstd{\mathbf{77.6}}{1.5} & \valstd{\mathbf{10.3}}{0.4} & \valstd{76.6}{3.6} & \valstd{\mathbf{10.9}}{1.1}    \\
			\bottomrule
		\end{tabular}%
	}
    \caption{\textbf{Inference-time comparison with baselines.} Success rates and planning times (sec.).}
	\label{tab:planning_time}
\end{table}
Table~\ref{tab:planning_time} compares planning time on the most challenging \texttt{PointMaze-Giant} and \texttt{AntMaze-Giant} tasks. MCDC improves over CompDiffuser by $9.6$ and $11.4$ success-rate points, respectively, while maintaining nearly the same planning time. Compared with CDGS, MCDC is about $5.6\times$ faster on \texttt{PointMaze-Giant} and $6.8\times$ faster on \texttt{AntMaze-Giant}. CDGS attains slightly higher success on \texttt{AntMaze-Giant}, but with substantially higher inference cost. Overall, MCDC provides a favorable success--cost trade-off by improving over CD at CD-level planning time and approaching CDGS-level success at much lower cost.

\paragraph{Forward evaluations and peak GPU memory.}
Table~\ref{tab:planning_nfe_memory} reports the forward evaluation count
and peak GPU memory on \texttt{PointMaze-Giant} and \texttt{AntMaze-Giant}.
CD and MCDC each use 100 denoiser forward evaluations per initial
candidate and trajectory chunk. CDGS and CDGS + MCDC each use 518.

\begin{table}[ht]
    \centering
    \begin{tabular}{@{}lrrr@{}}
        \toprule
        & & \multicolumn{2}{c@{}}{Peak GPU Memory (GiB)} \\
        \cmidrule(l){3-4}
        Method & Forward NFE & \texttt{PointMaze-Giant} & \texttt{AntMaze-Giant} \\
        \midrule
        CD          & 100 & \multirow{2}{*}{0.84} & \multirow{2}{*}{0.85} \\
        MCDC        & 100 & & \\
        \midrule
        CDGS        & 518 & \multirow{2}{*}{0.80} & \multirow{2}{*}{0.81} \\
        CDGS + MCDC & 518 & & \\
        \bottomrule
    \end{tabular}
    \caption{\textbf{Forward evaluations and peak GPU memory.}
    Forward NFE counts denoiser evaluations per initial candidate
    and per trajectory chunk. Peak memory is reported for each planner
    family in each environment.}
    \label{tab:planning_nfe_memory}
\end{table}

\subsection{Wall-validity Filtering.}
We compare CD and MCDC with and without a wall-validity filter on
\texttt{PointMaze-Giant} and \texttt{AntMaze-Giant}. Both filtered
variants use the same maze-map check to screen fused trajectories for
wall intersections, with $B=40$ candidates per planning call.

\begin{table}[ht]
    \centering
    \begin{tabular}{@{}lcc@{}}
        \toprule
        Method & \texttt{PointMaze-Giant} & \texttt{AntMaze-Giant} \\
        \midrule
        CD            & 68.0 & 65.2 \\
        CD + Filter   & 72.4 & 73.6 \\
        MCDC          & 77.6 & 76.6 \\
        MCDC + Filter & \textbf{81.0} & \textbf{84.8} \\
        \bottomrule
    \end{tabular}
    \caption{\textbf{Effect of wall-validity filtering on task success.}
    Values are mean episode success rates (\%) over five seeds.}
    \label{tab:wall_validity_filter}
\end{table}

Table~\ref{tab:wall_validity_filter} shows that filtering improved the
mean success rates of both planners on both mazes. Unfiltered MCDC
exceeded CD + Filter by 5.2 and 3.0 percentage points on
\texttt{PointMaze-Giant} and \texttt{AntMaze-Giant}, respectively.
With the same filter, MCDC exceeded CD by 8.6 and 11.2 percentage
points, respectively.

\subsection{Additional Comparisons with Related Planning Baselines}
\label{app:ecd_rcd_comparison}

We evaluate ECD~\cite{ECD}
and RCD~\cite{RCD} with and without
MCDC on four OGBench Stitch tasks. The combined variants apply
MCDC's coupled chunk-noise initialization to the respective planner.
We include standalone MCDC for reference.

\begin{table}[ht]
    \centering
    \begin{tabular}{@{}lccccc@{}}
        \toprule
        Dataset & ECD & RCD & MCDC & ECD + MCDC & RCD + MCDC \\
        \midrule
        \texttt{AntMaze-Large}
            & $79.0 \pm 2.0$ & $90.8 \pm 2.3$ & $96.0 \pm 0.7$ & $94.0 \pm 2.0$ & $96.4 \pm 1.8$ \\
        \texttt{AntMaze-Giant}
            & $80.6 \pm 3.3$ & $74.8 \pm 2.2$ & $76.6 \pm 3.6$ & $83.4 \pm 2.7$ & $87.6 \pm 2.7$ \\
        \texttt{HumanoidMaze-Large}
            & $57.2 \pm 3.7$ & $72.0 \pm 3.2$ & $79.8 \pm 3.3$ & $78.6 \pm 4.1$ & $80.2 \pm 3.3$ \\
        \texttt{HumanoidMaze-Giant}
            & $46.4 \pm 2.6$ & $53.4 \pm 4.2$ & $54.6 \pm 2.6$ & $56.2 \pm 1.9$ & $58.2 \pm 2.9$ \\
        \bottomrule
    \end{tabular}
    \caption{\textbf{Success rates on four OGBench Stitch tasks.}
    Values are mean $\pm$ standard deviation (\%) over five seeds.}
    \label{tab:ecd_rcd_comparison}
\end{table}

As shown in Table~\ref{tab:ecd_rcd_comparison}, adding MCDC increased
the mean success rate of each planner on all four tasks. Relative to
their respective baselines, the gains were 2.8--21.4 percentage points
for ECD and 4.8--12.8 percentage points for RCD.

\section{Visualizations}
\label{app:visualizations}

\subsection{Additional Mode Conflict Visualizations}
\label{app:mode_conflict_visualizations}

\begin{figure*}[h]
    \centering
    \includegraphics[width=0.95\linewidth]{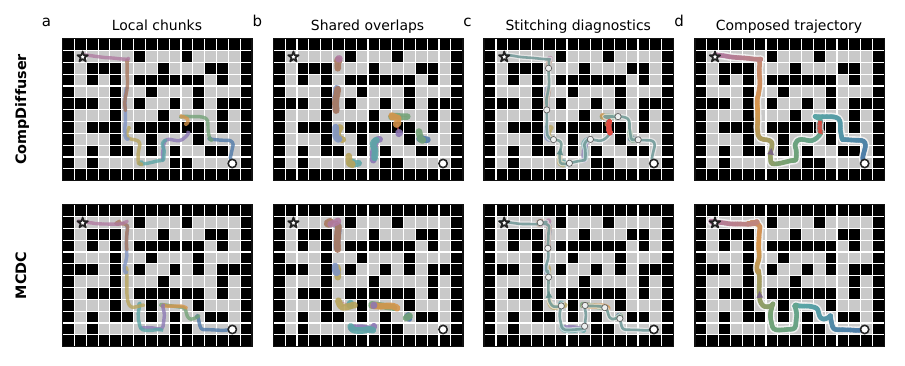}
    \caption{Mode conflict in \texttt{AntMaze-Giant-Stitch} task 2.}
    \label{fig:comparison_visualization_task2}
\end{figure*}

\begin{figure*}[h]
    \centering
    \includegraphics[width=0.95\linewidth]{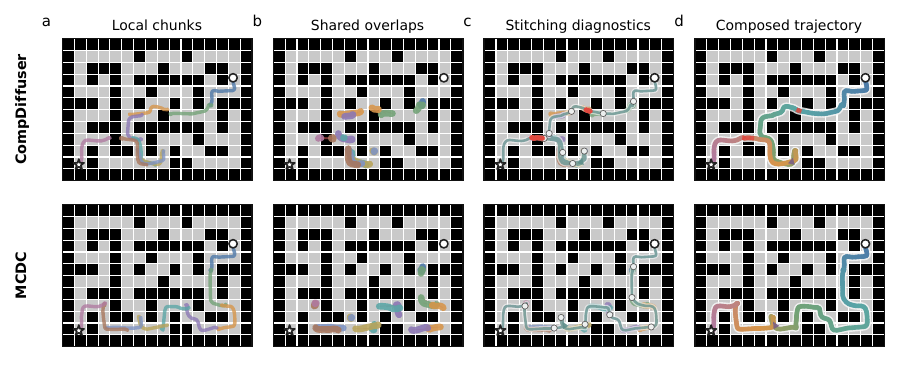}
    \caption{Mode conflict in \texttt{AntMaze-Giant-Stitch} task 3.}
    \label{fig:comparison_visualization_task3}
\end{figure*}

\begin{figure*}[h]
    \centering
    \includegraphics[width=0.95\linewidth]{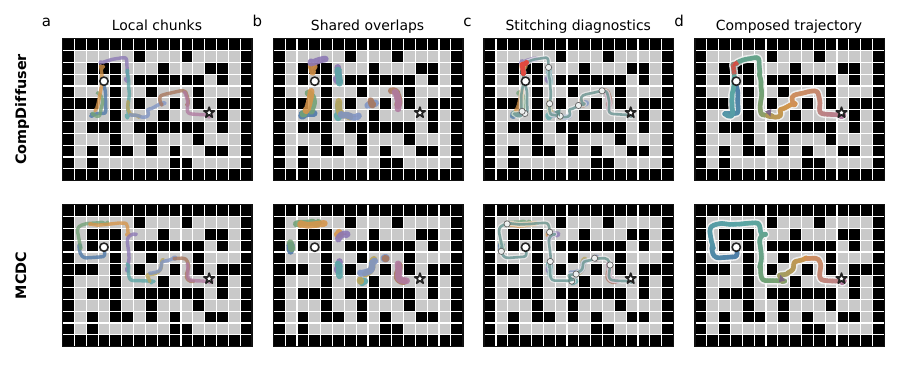}
    \caption{Mode conflict in \texttt{AntMaze-Giant-Stitch} task 4.}
    \label{fig:comparison_visualization_task4}
\end{figure*}

\begin{figure*}[h]
    \centering
    \includegraphics[width=0.95\linewidth]{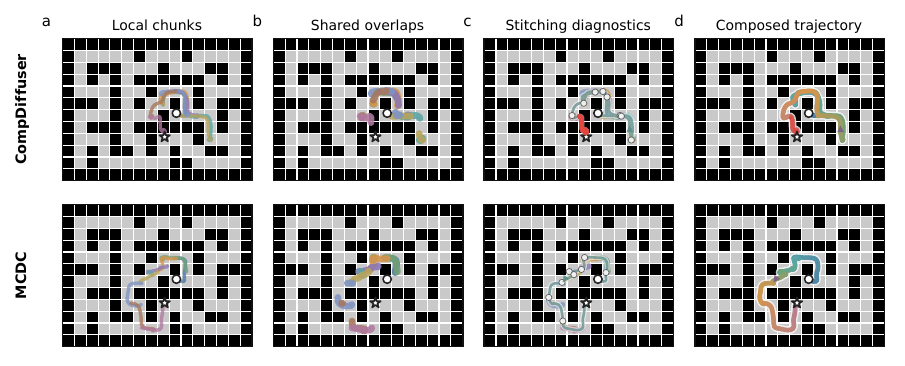}
    \caption{Mode conflict in \texttt{AntMaze-Giant-Stitch} task 5.}
    \label{fig:comparison_visualization_task5}
\end{figure*}

Figures~\ref{fig:comparison_visualization_task2}--\ref{fig:comparison_visualization_task5} provide additional examples on \texttt{AntMaze-Giant-Stitch}. They use the same diagnostic layout as Fig.~\ref{fig:comparison_visualization_task1}. Each figure shows local chunks, shared overlap regions, stitching diagnostics, and the final composed trajectory for CompDiffuser and MCDC. These examples illustrate that mode conflict can arise even when individual chunks are locally plausible: independently initialized chunks may select incompatible overlap modes, whereas MCDC biases neighboring chunks toward more compatible compositions.

\subsection{Planned Trajectory Visualizations}
\label{app:planed_trajectory}

\begin{figure*}[h]
    \centering
    \includegraphics[width=0.95\linewidth]{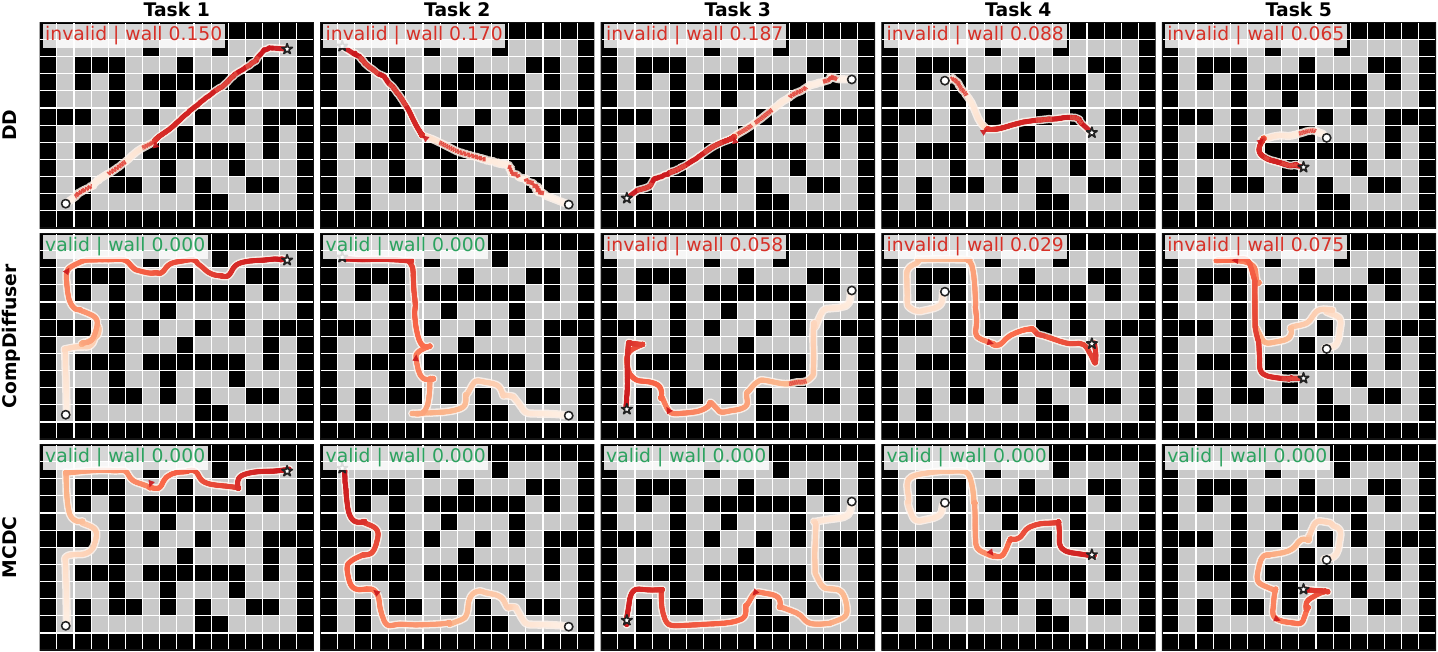}
    \caption{
    \textbf{Qualitative Comparison of DD, CompDiffuser and MCDC on \texttt{Pointmaze-Giant-Stitch}.} Each column shows one start-goal task, and each row shows the planned trajectory of one method. Green and red labels denote wall-valid and wall-invalid plans, respectively.
    }
    \label{fig:pointmaze_trajectory}
\end{figure*}

Figure~\ref{fig:pointmaze_trajectory} provides additional planned-trajectory visualizations on \texttt{Pointmaze-Giant-Stitch}. The figure complements Fig.~\ref{fig:antmaze_trajectory} by showing whether the composed plans respect maze geometry before execution. These qualitative examples support the same conclusion as the main text: MCDC improves the geometric feasibility of composed plans while using the same pretrained local diffusion model.

\subsection{Wall Validity of Initial Plans}
\label{app:first_plan_geometry}

We evaluate the first plan selected by each method before execution,
using an initial candidate batch of size $B=40$. For each method and
environment, we use the same five seeds, with 100 trials per seed.
We check the continuous planar centerlines of the original chunks and
the fused plan for wall intersections.

All three rates use the total number of selected first plans as their
denominator. \text{All Chunks Valid} is the fraction of plans with
every original chunk wall-valid. \text{Fusion-Induced Failures} is the
fraction with all original chunks wall-valid and a wall-invalid fused
plan. \text{Fused Plan Valid} is the fraction with a wall-valid fused
plan.

\begin{table}[ht]
    \centering
    \begin{tabular}{@{}llccc@{}}
        \toprule
        Dataset & Method
        & \text{All Chunks Valid (\%)}
        & \text{Fusion-Induced Failures (\%)}
        & \text{Fused Plan Valid (\%)} \\
        \midrule
        \multirow{3}{*}{\texttt{PointMaze-Giant}}
            & CD   & $73.4 \pm 2.6$ & $22.8 \pm 3.0$ & $51.2 \pm 3.8$ \\
            & CDGS & $56.0 \pm 3.6$ & $1.0 \pm 0.7$  & $55.0 \pm 3.4$ \\
            & MCDC & $90.4 \pm 1.3$ & $5.6 \pm 1.7$  & $84.8 \pm 2.6$ \\
        \midrule
        \multirow{3}{*}{\texttt{AntMaze-Giant}}
            & CD   & $36.4 \pm 2.1$ & $10.0 \pm 2.4$ & $28.0 \pm 4.6$ \\
            & CDGS & $87.6 \pm 4.6$ & $0.0 \pm 0.0$  & $87.6 \pm 4.6$ \\
            & MCDC & $85.8 \pm 3.6$ & $3.2 \pm 2.6$  & $85.2 \pm 3.2$ \\
        \bottomrule
    \end{tabular}
    \caption{\textbf{Wall validity of initial plans.}
    Rates are reported as mean $\pm$ standard deviation across five seeds.}
    \label{tab:first_plan_geometry}
\end{table}

Table~\ref{tab:first_plan_geometry} shows that MCDC achieved higher
fused-plan validity than CD on both mazes. Among first plans with
all chunks wall-valid, the pooled fusion failure rate ($L/A$) was
31.1\% for CD and 6.2\% for MCDC on \texttt{PointMaze-Giant},
and 27.5\% and 3.7\%, respectively, on \texttt{AntMaze-Giant}.
CDGS had the lowest fusion loss on both mazes and the highest mean
fused-plan validity on \texttt{AntMaze-Giant}.

%% file: reference.bib
@article{PCDT,
  title={PCDT: Pessimistic critic decision transformer for offline reinforcement learning},
  author={Wang, Xuesong and Zhang, Hengrui and Zhang, Jiazhi and Chen, CL Philip and Cheng, Yuhu},
  journal={IEEE Transactions on Systems, Man, and Cybernetics: Systems},
  volume={55},
  number={10},
  pages={7247--7258},
  year={2025},
  publisher={IEEE}
}

@article{DSP,
  title={Diffusion subgoal planning for long-horizon offline goal-conditioned reinforcement learning},
  author={Zhang, Hengrui and Cheng, Yuhu and Chen, CL and Wang, Xuesong},
  journal={arXiv preprint arXiv:2609.34575},
  year={2026}
}

@article{ECD,
  title={Energy-based Compositional Diffusion Planning},
  author={Sun, Tao and Mishra, Utkarsh Aashu and Lu, Jiaxin and Xu, Danfei and Armeni, Iro},
  journal={arXiv preprint arXiv:2606.21646},
  year={2026}
}

@inproceedings{TDP,
  title={Efficient Diffusion Planning with Temporal Diffusion},
  author={Guo, Jiaming and Zhang, Rui and Li, Zerun and Gao, Yunkai and Peng, Shaohui and Lan, Siming and Hu, Xing and Du, Zidong and Zhang, Xishan and Li, Ling},
  booktitle={Proceedings of the AAAI Conference on Artificial Intelligence},
  volume={40},
  number={26},
  pages={21450--21458},
  year={2026}
}

@article{PG,
  title={Prior-Guided Diffusion Planning for Offline Reinforcement Learning},
  author={Ki, Donghyeon and Oh, JunHyeok and Shim, Seong-Woong and Lee, Byung-Jun},
  journal={arXiv preprint arXiv:2505.10881},
  year={2025}
}

@article{GTS,
  title={Inference-time scaling of diffusion models through classical search},
  author={Zhang, Xiangcheng and Lin, Haowei and Ye, Haotian and Zou, James and Ma, Jianzhu and Liang, Yitao and Du, Yilun},
  journal={arXiv preprint arXiv:2505.23614},
  year={2025}
}

@article{FastMCTD,
  title={Fast monte carlo tree diffusion: 100$\times$ speedup via parallel and sparse planning},
  author={Yoon, Jaesik and Cho, Hyeonseo and Bengio, Yoshua and Ahn, Sungjin},
  journal={Advances in Neural Information Processing Systems},
  volume={38},
  pages={104526--104552},
  year={2026}
}

@inproceedings{MCTD,
  title={Monte Carlo Tree Diffusion for System 2 Planning},
  author={Yoon, Jaesik and Cho, Hyeonseo and Baek, Doojin and Bengio, Yoshua and Ahn, Sungjin},
  booktitle={International Conference on Machine Learning},
  pages={72618--72640},
  year={2025}
}

@article{LAVL,
  title={Latent Representation Alignment for Offline Goal-Conditioned Reinforcement Learning},
  author={Kang, Hyungkyu and Kim, Byeongchan and Oh, Min-hwan},
  journal={arXiv preprint arXiv:2605.25740},
  year={2026}
}

@article{SVL,
  title={SVL: Goal-Conditioned Reinforcement Learning as Survival Learning},
  author={Nguimatsia Tiofack, Franki and Schramm, Fabian and Le Hellard, Th{\'e}otime and Carpentier, Justin},
  journal={arXiv preprint arXiv:2604.17551},
  year={2026}
}

@article{CVP,
  title={Compositional visual planning via inference-time diffusion scaling},
  author={Zhang, Yixin and Luo, Yunhao and Mishra, Utkarsh Aashu and Shin, Woo Chul and Chen, Yongxin and Xu, Danfei},
  journal={arXiv preprint arXiv:2603.02646},
  year={2026}
}

@article{CinDM,
  title={Compositional generative inverse design},
  author={Wu, Tailin and Maruyama, Takashi and Wei, Long and Zhang, Tao and Du, Yilun and Iaccarino, Gianluca and Leskovec, Jure},
  journal={arXiv preprint arXiv:2401.13171},
  year={2024}
}

@inproceedings{CompositionalDiffusion,
  title={Compositional visual generation with composable diffusion models},
  author={Liu, Nan and Li, Shuang and Du, Yilun and Torralba, Antonio and Tenenbaum, Joshua B},
  booktitle={European conference on computer vision},
  pages={423--439},
  year={2022},
  organization={Springer}
}

@article{AdaptDiffuser,
  title={Adaptdiffuser: Diffusion models as adaptive self-evolving planners},
  author={Liang, Zhixuan and Mu, Yao and Ding, Mingyu and Ni, Fei and Tomizuka, Masayoshi and Luo, Ping},
  journal={arXiv preprint arXiv:2302.01877},
  year={2023}
}

@inproceedings{GSC,
  title={Generative skill chaining: Long-horizon skill planning with diffusion models},
  author={Mishra, Utkarsh Aashu and Xue, Shangjie and Chen, Yongxin and Xu, Danfei},
  booktitle={Conference on Robot Learning},
  pages={2905--2925},
  year={2023},
  organization={PMLR}
}

@article{XDiffuser,
  title={Plan First, Diffuse Later: Extrinsic Graph Guidance for Long-Horizon Diffusion Planning},
  author={Hassidof, Yaniv and Morgan, Adir and Du, Yilun and Solovey, Kiril},
  journal={arXiv preprint arXiv:2605.16863},
  year={2026}
}

@article{CDGS,
  title={Compositional Diffusion with Guided search for Long-Horizon Planning},
  author={Mishra, Utkarsh A and He, David and Chen, Yongxin and Xu, Danfei},
  journal={arXiv preprint arXiv:2601.00126},
  year={2025}
}

@article{RCD,
  title={Refining Compositional Diffusion for Reliable Long-Horizon Planning},
  author={Lee, Kyowoon and Luo, Yunhao and Tong, Anh and Choi, Jaesik},
  journal={arXiv preprint arXiv:2605.03075},
  year={2026}
}

@article{CompDiffuser,
  title={Generative trajectory stitching through diffusion composition},
  author={Luo, Yunhao and Mishra, Utkarsh and Du, Yilun and Xu, Danfei},
  journal={Advances in Neural Information Processing Systems},
  volume={38},
  pages={37809--37843},
  year={2025}
}

@article{FQL,
  title={Flow Q-Learning},
  author={Seohong Park and Qiyang Li and Sergey Levine},
  journal={arXiv preprint arXiv:2502.02538},
  year={2025},
}

@article{DTAMP,
  title={Diffused Task-Agnostic Milestone Planner},
  author={Hong, Mineui and Kang, Minjae and Oh, Songhwai},
  journal={Advances in Neural Information Processing Systems},
  volume={36},
  pages={387--405},
  year={2023}
}

@inproceedings{HDMI,
  title={Hierarchical diffusion for offline decision making},
  author={Li, Wenhao and Wang, Xiangfeng and Jin, Bo and Zha, Hongyuan},
  booktitle={International Conference on Machine Learning},
  pages={20035--20064},
  year={2023},
  organization={PMLR}
}

@article{HD,
  title={Simple hierarchical planning with diffusion},
  author={Chen, Chang and Deng, Fei and Kawaguchi, Kenji and Gulcehre, Caglar and Ahn, Sungjin},
  journal={arXiv preprint arXiv:2401.02644},
  year={2024}
}

@article{CFGRL,
  title={Diffusion Guidance Is a Controllable Policy Improvement Operator},
  author={Frans, Kevin and Park, Seohong and Abbeel, Pieter and Levine, Sergey},
  journal={arXiv preprint arXiv:2505.23458},
  year={2025}
}

@article{DiffusionQL,
  title={Diffusion policies as an expressive policy class for offline reinforcement learning},
  author={Wang, Zhendong and Hunt, Jonathan J and Zhou, Mingyuan},
  journal={arXiv preprint arXiv:2208.06193},
  year={2022}
}

@article{FM,
  title={Flow matching for generative modeling},
  author={Lipman, Yaron and Chen, Ricky TQ and Ben-Hamu, Heli and Nickel, Maximilian and Le, Matt},
  journal={arXiv preprint arXiv:2210.02747},
  year={2022}
}

@article{RF,
  title={Flow straight and fast: Learning to generate and transfer data with rectified flow},
  author={Liu, Xingchao and Gong, Chengyue and Liu, Qiang},
  journal={arXiv preprint arXiv:2209.03003},
  year={2022}
}

@article{DDIM,
  title={Denoising diffusion implicit models},
  author={Song, Jiaming and Meng, Chenlin and Ermon, Stefano},
  journal={arXiv preprint arXiv:2010.02502},
  year={2020}
}

@article{DDPM,
  title={Denoising diffusion probabilistic models},
  author={Ho, Jonathan and Jain, Ajay and Abbeel, Pieter},
  journal={Advances in Neural Information Processing Systems},
  volume={33},
  pages={6840--6851},
  year={2020}
}

@article{Score_SDE,
  title={Score-based generative modeling through stochastic differential equations},
  author={Song, Yang and Sohl-Dickstein, Jascha and Kingma, Diederik P and Kumar, Abhishek and Ermon, Stefano and Poole, Ben},
  journal={arXiv preprint arXiv:2011.13456},
  year={2020}
}

@article{Pi-HIQL,
  title={Physics-informed Value Learner for Offline Goal-Conditioned Reinforcement Learning},
  author={Giammarino, Vittorio and Ni, Ruiqi and Qureshi, Ahmed H},
  journal={arXiv preprint arXiv:2509.06782},
  year={2025}
}

@article{Diffuser,
  title={Planning with diffusion for flexible behavior synthesis},
  author={Janner, Michael and Du, Yilun and Tenenbaum, Joshua B and Levine, Sergey},
  journal={arXiv preprint arXiv:2205.09991},
  year={2022}
}

@article{DD,
  title={Is conditional generative modeling all you need for decision-making?},
  author={Ajay, Anurag and Du, Yilun and Gupta, Abhi and Tenenbaum, Joshua and Jaakkola, Tommi and Agrawal, Pulkit},
  journal={arXiv preprint arXiv:2211.15657},
  year={2022}
}

@inproceedings{SkillDiffuser,
  title={Skilldiffuser: interpretable hierarchical planning via skill abstractions in diffusion-based task execution},
  author={Liang, Zhixuan and Mu, Yao and Ma, Hengbo and Tomizuka, Masayoshi and Ding, Mingyu and Luo, Ping},
  booktitle={Proceedings of the IEEE/CVF Conference on Computer Vision and Pattern Recognition},
  pages={16467--16476},
  year={2024}
}

@article{CHER,
  title={Curriculum-guided hindsight experience replay},
  author={Fang, Meng and Zhou, Tianyi and Du, Yali and Han, Lei and Zhang, Zhengyou},
  journal={Advances in Neural Information Processing Systems},
  volume={32},
  year={2019}
}

@article{HER,
  title={Hindsight experience replay},
  author={Andrychowicz, Marcin and Wolski, Filip and Ray, Alex and Schneider, Jonas and Fong, Rachel and Welinder, Peter and McGrew, Bob and Tobin, Josh and Pieter Abbeel, OpenAI and Zaremba, Wojciech},
  journal={Advances in Neural Information Processing Systems},
  volume={30},
  year={2017}
}

@article{OGBench,
  title={Ogbench: benchmarking offline goal-conditioned rl},
  author={Park, Seohong and Frans, Kevin and Eysenbach, Benjamin and Levine, Sergey},
  journal={arXiv preprint arXiv:2410.20092},
  year={2024}
}

@article{IQL,
  title={Offline reinforcement learning with implicit q-learning},
  author={Kostrikov, Ilya and Nair, Ashvin and Levine, Sergey},
  journal={arXiv preprint arXiv:2110.06169},
  year={2021}
}

@article{HIQL,
  title={Hiql: offline goal-conditioned rl with latent states as actions},
  author={Park, Seohong and Ghosh, Dibya and Eysenbach, Benjamin and Levine, Sergey},
  journal={Advances in Neural Information Processing Systems},
  volume={36},
  pages={34866--34891},
  year={2023}
}

@inproceedings{QRL,
  title={Optimal goal-reaching reinforcement learning via quasimetric learning},
  author={Wang, Tongzhou and Torralba, Antonio and Isola, Phillip and Zhang, Amy},
  booktitle={International Conference on Machine Learning},
  pages={36411--36430},
  year={2023},
  organization={PMLR}
}

@article{CRL,
  title={Contrastive learning as goal-conditioned reinforcement learning},
  author={Eysenbach, Benjamin and Zhang, Tianjun and Levine, Sergey and Salakhutdinov, Russ R},
  journal={Advances in Neural Information Processing Systems},
  volume={35},
  pages={35603--35620},
  year={2022}
}

@article{OTA,
  title={Option-aware temporally abstracted value for offline goal-conditioned reinforcement learning},
  author={Ahn, Hongjoon and Choi, Heewoong and Han, Jisu and Moon, Taesup},
  journal={arXiv preprint arXiv:2505.12737},
  year={2025}
}

@article{SAW,
  title={Flattening hierarchies with policy bootstrapping},
  author={Zhou, John and Kao, Jonathan},
  journal={Advances in Neural Information Processing Systems},
  volume={38},
  pages={82975--83007},
  year={2026}
}

@article{GoFAR,
  title={Offline goal-conditioned reinforcement learning via $ f $-advantage regression},
  author={Ma, Jason Yecheng and Yan, Jason and Jayaraman, Dinesh and Bastani, Osbert},
  journal={Advances in Neural Information Processing Systems},
  volume={35},
  pages={310--323},
  year={2022}
}

@article{GCSL,
  title={Learning to reach goals via iterated supervised learning},
  author={Ghosh, Dibya and Gupta, Abhishek and Reddy, Ashwin and Fu, Justin and Devin, Coline and Eysenbach, Benjamin and Levine, Sergey},
  journal={arXiv preprint arXiv:1912.06088},
  year={2019}
}

@article{WGCSL,
  title={Rethinking goal-conditioned supervised learning and its connection to offline rl},
  author={Yang, Rui and Lu, Yiming and Li, Wenzhe and Sun, Hao and Fang, Meng and Du, Yali and Li, Xiu and Han, Lei and Zhang, Chongjie},
  journal={arXiv preprint arXiv:2202.04478},
  year={2022}
}

@article{Goplan,
  title={Goplan: goal-conditioned offline reinforcement learning by planning with learned models},
  author={Wang, Mianchu and Yang, Rui and Chen, Xi and Sun, Hao and Fang, Meng and Montana, Giovanni},
  journal={arXiv preprint arXiv:2310.20025},
  year={2023}
}
